\documentclass[runningheads]{llncs}
\usepackage{amsmath,amsfonts,bm}

\def\1{\bm{1}}

\newcommand{\test}{\mathcal{D_{\mathrm{test}}}}

\def\vf{{\bm{f}}}

\def\mA{{\bm{A}}}
\def\mB{{\bm{B}}}

\def\mF{{\bm{F}}}

\def\mI{{\bm{I}}}

\def\mP{{\bm{P}}}
\def\mQ{{\bm{Q}}}
\def\mR{{\bm{R}}}
\def\mS{{\bm{S}}}

\def\mW{{\bm{W}}}
\def\mX{{\bm{X}}}

\def\mZ{{\bm{Z}}}

\def\mLambda{{\bm{\Lambda}}}
\def\mSigma{{\bm{\Sigma}}}

\DeclareMathAlphabet{\mathsfit}{\encodingdefault}{\sfdefault}{m}{sl}
\SetMathAlphabet{\mathsfit}{bold}{\encodingdefault}{\sfdefault}{bx}{n}

\def\gP{{\mathcal{P}}}

\def\gR{{\mathcal{R}}}

\def\emC{{C}}

\newcommand{\R}{\mathbb{R}}

\usepackage{graphicx}
\usepackage{float, subfig}
\usepackage{comment}
\usepackage{amsmath,amssymb} % define this before the line numbering.
\usepackage{color}
\usepackage{booktabs, multirow}
\usepackage{makecell}
\usepackage{grffile}

\usepackage[pagebackref=true,
            breaklinks=true,
            letterpaper=true,
            colorlinks,
            bookmarks=false,
            citecolor=blue,
            linkcolor=blue]{hyperref}

\graphicspath{{./extra/}}
\DeclareGraphicsExtensions{.pdf,.jpg,.png}

\newcolumntype{L}[1]{>{\raggedright\arraybackslash}p{#1}}
\newcolumntype{C}[1]{>{\centering\arraybackslash}p{#1}}
\newcolumntype{R}[1]{>{\raggedleft\arraybackslash}p{#1}}

\def\SO{\mathrm{SO}}
\def\diag{\mathrm{diag}}
\def\skew{\mathrm{skew}}

\def\etal{{\em et al.~}}

\def\ie{{\em i.e.,~}}

\newtheorem{thm}{Theorem}
\newtheorem{defn}{Definition}

\begin{document}
% \linenumbers
\pagestyle{headings}
\mainmatter
\def\ECCVSubNumber{872}  % Insert your submission number here

\title{Learnable Cost Volume Using the Cayley Representation} % Replace with your title

% \begin{comment}
% \titlerunning{ECCV-20 submission ID \ECCVSubNumber} 
% \authorrunning{ECCV-20 submission ID \ECCVSubNumber} 
% \author{Anonymous ECCV submission}
% \institute{Paper ID \ECCVSubNumber}
% \end{comment}
%******************

% \begin{comment}
\titlerunning{Learnable Cost Volume Using the Cayley Representation}
\author{Taihong Xiao\inst{1}\orcidID{0000-0002-6953-7100} \and
Jinwei Yuan\inst{2} \and
Deqing Sun\inst{2}\orcidID{0000-0003-0329-0456} \and
Qifei Wang\inst{2}
Xin-Yu Zhang\inst{3} \and
Kehan Xu\inst{4} \and
Ming-Hsuan Yang\inst{1,2}\orcidID{0000-0003-4848-2304}
}
\authorrunning{T. Xiao et al.}
\institute{University of California, Merced \email{\{txiao3,mhyang\}@ucmerced.edu} \and
Google Research \email{\{jinwei,deqingsun,qfwang,minghsuan\}@google.com} \and
Nankai University \email{xinyuzhang@mail.nankai.edu.cn} \and
Peking University \email{yurina@pku.edu.cn}
% 
% \institute{University of California, Merced \and
% Google Research \and
% Nankai University \and
% Peking University 
}
% \end{comment}
%******************
\maketitle

\begin{abstract}
Cost volume is an essential component of recent deep models for optical flow estimation
and is usually constructed by calculating the inner product between two feature vectors.
However, the standard inner product in the commonly-used cost volume
may limit the representation capacity of flow models
because it neglects the correlation among different channel dimensions
and weighs each dimension equally.
To address this issue, we propose a \emph{learnable cost volume} (LCV) using an elliptical inner product, which generalizes the standard inner product by a positive definite kernel matrix.
To guarantee its positive definiteness, we perform spectral decomposition on the kernel matrix and re-parameterize it via the Cayley representation.
The proposed LCV is a lightweight module and can be easily plugged into existing models to replace the vanilla cost volume.
Experimental results show that the LCV module not only improves the accuracy of state-of-the-art models on standard benchmarks, but also promotes their robustness against illumination change,
noises, and adversarial perturbations of the input signals.
\end{abstract}

% \Uspace
\section{Introduction}\label{sec:introduction}
% \Lspace

Optical flow estimation is a fundamental computer vision task and
has broad applications, such as video interpolation~\cite{DAIN},
video prediction~\cite{Prediction-ECCV-2018},
video segmentation~\cite{tsai2016video,cheng2017segflow},
and action recognition~\cite{lin2019tsm}.
Despite the recent progress made by deep learning models,
it is still challenging to accurately estimate optical flow
for image sequences with large displacements, textureless regions, motion blur, occlusion,
illumination changes, and non-Lambertian reflection.

\begin{figure}[!htbp]
    \centering
    \includegraphics[width=0.6\textwidth]{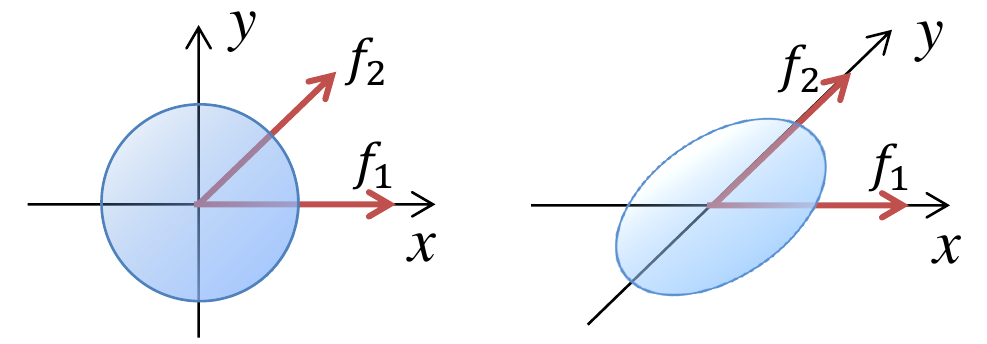}
    \caption{Standard inner product space v.s. elliptical inner product space.}
    \label{fig:inner_product}
\end{figure}

Most deep optical flow models~\cite{sun2018pwc,Liu:2019:DDFlow,hui2018liteflownet}
adopt the idea of coarse-to-fine processing via feature pyramids and
construct {\it cost volumes} at different levels of the pyramids.
The cost volume stores the costs of matching pixels in the source image
with their potential matching candidates in the target image. 
It is typically constructed by calculating the inner product between
the convolutional features of one frame and those of the next frame,
and then regressed to the estimated optical flow by an estimation sub-network.
The accuracy of the estimated optical flow heavily relies on the quality of the constructed cost volume.

While the standard Euclidean inner product is widely used to build the cost volume (a.k.a., vanilla cost volume) for optical flow, we argue that it limits the representation capacity of the flow model for two reasons.
First, the correlation among different channel dimensions is not taken
into consideration by the standard Euclidean inner product.
As shown in Fig.~\ref{fig:inner_product}, we use a simple 2D example for illustration.
Given two feature vectors $\vf_1$ and $\vf_2$ with positive correlation
in the standard inner product space, we are able to find a proper
elliptical inner product space to make these two feature vectors
orthogonal to each other, which gives a zero correlation.
Therefore, the specific choice of the inner product space influences
the values of the matching costs, and thus should be further exploited.
Second, each feature dimension contributes equally to the vanilla cost volume,
which may give a sub-optimal solution to constructing the cost volume
for flow estimation.
Ideally, dimensions corresponding to noises and random perturbations should be suppressed, while those containing discriminative signals for flow estimation should be kept or magnified. 

To address these limitations, we propose a \textit{learnable cost volume} (LCV) module
which accounts for the correlation among different channel dimensions
and re-weighs the contribution of each feature channel to the cost volume.
The LCV generalizes the Euclidean inner product space to an elliptical
inner product space, which is parameterized by a symmetric and
positive definite kernel matrix.
The spectral decomposition of the kernel matrix gives an orthogonal matrix
and a diagonal matrix.
The orthogonal matrix linearly transforms the features into a new feature space,
which accounts for the correlation among different channel dimensions.
The diagonal matrix multiplies each transformed feature by a positive scalar,
which weighs each feature dimension differently.
From a geometric perspective, the orthogonal matrix rotates the axes and
the diagonal matrix stretches the axes so that the feature vectors are represented in a learned elliptical inner product space, which generates more discriminative matching costs for flow estimation.

However, directly learning a kernel matrix in an end-to-end manner cannot
guarantee the symmetry and positive definiteness of the kernel matrix,
which is required by the definition of inner product.
To address this issue, we perform spectral decomposition on
the kernel matrix and represent each component via the Cayley transform.
Specifically, the special orthogonal matrices that exclude $-1$ as the eigenvalue
can be bijectively mapped into the skew-symmetric matrices,
and the diagonal matrices can be similarly represented by the composition
of the Cayley transform and the arctangent function.
In this way, all parameters of the learnable cost volume can be inferred
in an end-to-end fashion without explicitly imposing any constraints.

The proposed learnable cost volume is a general version of the vanilla cost volume,
and thus can replace the vanilla cost volume in the existing networks.
We finetune the existing architectures equipped with LCV
by initializing the kernel matrix as the identity matrix
and restoring other parameters from the pre-trained models.
Experimental results on the Sintel and KITTI benchmark datasets show that
the proposed LCV significantly improves the performance of existing methods
in both supervised and unsupervised settings.
In addition, we demonstrate that LCV is able to promote the robustness
of the existing models against illumination changes, noises,
and adversarial attacks.

To summarize, we make the following contributions:
\begin{enumerate}
    \item We propose a learnable cost volume (LCV) 
    to account for correlations among different feature dimensions
    and weight each dimension separately.
    
    \item We employ the Cayley representation to
    re-parameterize the kernel matrix in a way
    that all parameters can be learned in an end-to-end manner.
    
    \item The proposed LCV can easily replace the vanilla cost volume
    and improve the accuracy and robustness of the state-of-the-art models.
\end{enumerate}

% \Uspace
\section{Related Work}

% \PARAspace
\subsubsection{Supervised Learning of Optical Flow.}
Inspired by the success of convolutional neural networks (CNNs) on per-pixel predictions
such as semantic segmentation and single-image depth estimation,
Dosovitski~\etal propose FlowNet~\cite{dosovitskiy2015flownet}, the first end-to-end
deep neural network capable of learning optical flow.
FlowNet predicts a dense optical flow map from two consecutive image frames
with an encoder-decoder architecture.
FlowNet2.0~\cite{ilg2017flownet} extends FlowNet by stacking multiple basic FlowNet modules
for iterative refinement and its accuracy is fully on par with those of the
state-of-the-art methods at the time.
Motivated by the idea of coarse-to-fine refinement in traditional optical flow methods,
SpyNet~\cite{ranjan2017optical} introduces a compact spatial pyramid network
that warps images at multiple scales to deal with displacements caused by large motions.
PWC-Net~\cite{sun2018pwc} extracts feature through pyramidal processing and
builds a cost volume at each level from the warped and the target features
to iteratively refine the estimated flow.
VCN~\cite{yang2019vcn} improves the cost volume processing by decoupling
the 4D convolution into a 2D spatial filter and a 2D winner-take-all (WTA) filter,
while still retaining a large receptive field.
HD$^3$~\cite{Yin_2019_CVPR} learns a probabilistic matching density distribution
at each scale and merges the matching densities at different scales
to recover the global matching density. 

% \PARAspace
\subsubsection{Unsupervised Learning of Optical Flow.}
The advantage of unsupervised methods is that it can sidestep the limitations
of the synthetic datasets and exploit the large number of training data in the
realistic domain.
In \cite{jason2016back} and \cite{ren2017unsupervised}, the flow guidance comes from
warping the target image according to the predicted flow and comparing
against the reference image.
The photometric loss is adopted to ensure brightness constancy and
spatial smoothness.
In some work~\cite{wang2018occlusion,meister2018unflow}, occluded regions
are excluded from the photometric loss.
As pixels occluded in the target image are also absent in the warped one,
enforcing matching of the occluded pixels would misguide the training.
Wang~\etal~\cite{wang2018occlusion} obtain an occlusion mask from the range map
inferred from the backward flow,
while UnFlow~\cite{meister2018unflow} relies on the forward-backward consistency
to estimate the occlusion mask.
Unlike these two methods that predict the occlusion map in advance with certain heuristic,
Back2Future~\cite{janai2018unsupervised} estimates the occlusion and optical flow jointly
by introducing a multi-frame formulation and
reasoning the occlusion in a more advanced manner.
DDFlow~\cite{Liu:2019:DDFlow} performs knowledge distillation
by cropping patches from the unlabeled images,
which provides flow guidance for the occluded regions.
SelFlow~\cite{Liu:2019:SelFlow} hallucinates synthetic occlusions
by perturbing super-pixels where the occluded regions are guided by
a model pre-trained from non-occluded regions.

% \PARAspace
\subsubsection{Correspondence Matching.}
Typically, stereo matching algorithms~\cite{scharstein2002taxonomy,horn1981determining} involve local correspondence extraction and smoothness regularization,
where the smoothness regularization is enforced by energy minimization.
Recently, hand-crafted features are replaced by deep features and
minimization of the matching cost is substituted by training convolutional neural networks~\cite{zbontar2016stereo,kendall2017end}. 
Xu~\etal~\cite{XRK2017} construct a 4D cost volume using an adaptation of
the semi-global matching,
and Yang~\etal\cite{yang2019vcn} reduce the computation overhead of processing
the 4D matching volume by factorizing into two separable filters.

Different from these approaches where the correspondence is represented
by a hand-crafted matching cost volume, we propose a learnable cost volume
that can capture the correlation among different channels by adapting the features to an elliptical inner product space.
Such a correlation is automatically learned by optimizing the kernel matrix using the Cayley representation,
which is more flexible and effective in optical flow estimation and can be easily plugged into the existing architectures. To our knowledge, this paper is the first one to use the Cayley representation for learning correspondence in optical flow.

% \Uspace
\section{Learnable Correlation Volume}
% \Lspace

\subsection{Vanilla Cost Volume}
% \Lspace
Let $\mF^1, \mF^2 \in \R^{c \times h \times w}$ be the convolutional feature
of the first frame and the warped feature of the second frame, respectively.
The vanilla cost volume is defined as the inner product between
the query feature $\mF^{1}_{i,j}$ and the potential match candidate $\mF^{2}_{k',l'}$, \ie
\begin{equation}\label{eq:cost_volume}
    \emC(\mF^{1}, \mF^{2})_{k,l,i,j} = \mF^{1\top}_{i,j} \mF^{2}_{k',l'},
\end{equation}
which maps from the space $\R^{c \times h \times w}\times \R^{c \times h \times w}$ to $\R^{u\times v\times h\times w}$.
Here, $u$ and $v$ are usually odd numbers, indicating the displacement ranges in horizontal and vertical directions,
$(i,j)$ denotes the spatial location of the feature map $\mF^{1}$,
and $(k',l')=(i-(u-1)/2+k, j-(v-1)/2+l)$ denotes that of $\mF^{2}$.
For each location $(i,j)$ of the query feature $\mF^{1}$, the matching is performed against
pixels of $\mF^{2}$ within a $u \times v$ search window centered by the location $(i,j)$. 
Then, the cost volume is either reshaped into $uv \times h \times w$
and post-processed by 2D convolutions~\cite{sun2018pwc},
or kept as a 4D tensor on which the separable 4D convolutions~\cite{yang2019vcn} are applied.

% \Uspace
\subsection{Learnable Cost Volume}
% \Lspace
We generalize the standard Euclidean inner product to the elliptical inner product, where the matching cost is computed as follows:
\begin{equation}\label{eq:learnable_correlation_volume}
    \emC(\mF^{1}, \mF^{2})_{k,l,i,j} = \mF^{1\top}_{i,j} \mW \mF^{2}_{k',l'}.
\end{equation}
Here, $\mW \in \R^{c \times c}$ is a learnable kernel matrix that determines
the elliptical inner product space,
and other notations are the same as those in Eq.~\eqref{eq:cost_volume}.
According to the definition of inner product, $\mW$ should be a symmetric and positive definite matrix.
By spectral decomposition, we obtain
\begin{equation}\label{eq:spectral_decomposition}
    \mW = \mP^{\top} \mLambda \mP,
\end{equation}
where $\mP$ is an orthogonal matrix, and $\mLambda$ is a diagonal matrix
with positive entries, \ie $\mLambda = \diag(\lambda_1,\cdots,\lambda_c)$
with $\lambda_i > 0,~\forall i \in \{1,\cdots,c\}$.
The orthogonal matrix $\mP$ actually rotates the coordinate axes and
the diagonal matrix $\mLambda$ re-weights different dimensions,
which directly address the two limitations mentioned in Sec.~\ref{sec:introduction}.

% \Uspace
\subsection{Learning with the Cayley Representation}
% \Lspace

In the proposed LCV module, the entries of the kernel matrix $\mW$ are
the only learnable parameters.
However, the constraints of symmetry and positive-definiteness
hinders the gradient-based end-to-end learning of $\mW$.
To address this issue, we propose to optimize $\mP$ and $\mLambda$ instead of $\mW$.

One way to optimize $\mP$ is to employ the Riemann gradient descent
on the Stiefel manifold, which is defined as
\begin{equation}
    V_k(\R^n) = \{\mA\in\R^{n\times k} | \mA^\top \mA = \mI_k\}.
\end{equation}
All orthogonal matrices lie in the Stiefel manifold.
Specifically, $\mP \in V_c(\R^c)$.
Therefore, we can apply the Riemann gradient descent on the Stiefel matrix manifold,
where the projection and retraction formula~\cite{absil2009optimization} are given by
\begin{align}\label{eq:projection_and_rectraction}
	\gP_{\mX}(\mZ) &= (\mI-\mX\mX^\top)\mZ + \mX\cdot\skew(\mX^\top \mZ)\\
	\gR_{\mX}(\mZ) &= (\mX+\mZ)(\mI+\mZ^\top \mZ)^{-\frac{1}{2}},
\end{align}
where $\skew(\mX) := (\mX - \mX^\top)/2$.
However, to perform the Riemann gradient descent,
the projection and retraction operations are required in each training step,
and the matrix multiplication brings considerable computational overhead.

We can address this issue in a more elegant way using the Cayley Representation~\cite{cayley1846algebraic}.
First, we define a set of matrices:
\begin{align}
    \SO^*(n) := \{\mA \in \SO(n) : -1\not\in \sigma(\mA) \},
\end{align}
where $\sigma(\mA)$ denotes the spectrum, \ie all eigenvalues, of $\mA$.
$\SO^*(n)$ is a subset of the special orthogonal group $\SO(n)$ and the spectrum
of its elements excludes $-1$.
Then, we have the following theorems:
\begin{thm}[Cayley Representation]\label{thm:cayley}
Given any matrix $\mP \in \SO^*(n)$, there exists a unique skew-symmetric matrix $\mS$,
\ie $\mS^\top=-\mS$, such that
\begin{align}
    \mP = (\mI - \mS)(\mI + \mS)^{-1}.
\end{align}
\end{thm}

\begin{thm}\label{thm:connected}
    The set of matrices $\SO^*(n)$ is connected.
\end{thm}
% \ds{explain/define what connected mean.}

By Theorem~\ref{thm:cayley}, we can initialize the matrix $\mP$
in Eq.~\eqref{eq:spectral_decomposition} as an identity matrix $\mI\in\SO^*(c)$,
and update $\mS$ so as to update $\mP$ using gradient-based optimizer.
Let $\mP^*$ be the optimal orthogonal matrix, and we claim that it is possible
to reach $\mP^{*}$ from initializing as the identity matrix $\mP=\mI$.
This because $\SO^{*}(c)$ is a connected set (Theorem~\ref{thm:connected}),
so there exists a continuous path joining $\mI \in \SO^{*}(c)$ and
any $\mP \in \SO^{*}(c)$, including $\mP^{*}$.

Due to the positive definiteness of $\mW$, the constraint of the diagonal matrix
$\mLambda=\diag(\lambda_1,\ldots,\lambda_c)$ is $\lambda_i >0,~\forall i=1,\ldots,c$.
Thus, we map $\R$ to $\R^+$ by applying the composition of the Cayley transform
and the arctangent function, \ie
\begin{equation}
    \lambda_i = \frac{\pi + 2 \arctan{t_{i}}}{\pi - 2 \arctan{t_{i}}},
\end{equation}
where $t_{i} \in \R$ is free of constraint.

The above re-parameterization trick enables us to update the kernel matrix $\mW$
in an end-to-end manner using the SGD optimizer or its variants, which alleviates the heavy computation brought by the projection and retraction and makes the training process much easier.

\subsection{Interpretation}
\label{sec:interpretation}

To better understand the learnable cost volume,
we analyze several cases here.

\subsubsection{1. $\mW = \mI$.}
This degenerates into the vanilla cost volume, in which the standard Euclidean
inner product is adopted.

\subsubsection{2. $\mW = \mSigma^{-1}$.}
Let $\mSigma$ be the covariance matrix, \ie Gram matrix, of the convolutional feature, then the learnable cost volume is essentially a whitening transformation.
Let $\mQ = \mLambda^{1/2}\mP$, and then Eq.~\eqref{eq:learnable_correlation_volume} can be
formulated as
\begin{equation}\label{eq:PCA}
    \emC(\mF^{1}, \mF^{2})_{k,l,i,j} = \mF^{1\top}_{i,j} \mP^{\top} \mLambda^{1/2} \mLambda^{1/2} \mP \mF^{2}_{k',l'} = (\mQ \mF^{1}_{i,j})^{\top} (\mQ \mF^{2}_{k',l'}),
\end{equation}
where $\mQ \mF^1_{i,j}$ represents the transformed feature of $\mF^1_{i,j}$
after PCA~\cite{jolliffe1986principal} whitening.
Similarly, letting $\mR = \mP^{\top} \mLambda^{1/2}\mP$, we can have
\begin{equation}\label{eq:ZCA}
    \emC(\mF^{1}, \mF^{2})_{k,l,i,j} = \mF^{1\top}_{i,j} \mP^{\top} \mLambda^{1/2} \mP \mP^{\top} \mLambda^{1/2} \mP \mF^{2}_{k',l'} = (\mR \mF^{1}_{i,j})^{\top} (\mR \mF^{2}_{k',l'}),
\end{equation}
where $\mR \mF^1_{i,j}$ is the transformed feature of $\mF^1_{i,j}$
after ZCA~\cite{brox2010large} whitening.
It has been shown that the high-level styles can be removed with the contextual structures
remained by whitening the convolutional features~\cite{li2017universal}.

% \PARAspace
\subsubsection{3. $\mW = \mP^{\top} \mLambda \mP$.} The learnable cost volume shares a similar formula as the whitening process, but $\mW$ is learned over the whole training dataset rather than statistics of two inputs, thus contains certain holistic information of the entire training dataset. Because it has been verified that the certain holistic characteristics of the underlying image can be captured by the Gram matrix along the channel dimension \cite{gatys2016image,li2017universal}. The learnable cost volume performs as “whitening” features using the common information learned from all frames. Specifically, the orthogonal matrix $\mP$ re-arranges the information across the channel dimension, while the diagonal matrix $\mLambda$ filters out insignificant signals, making the correlation more robust to the illumination changes abd noises.~(See Sec.~\ref{sec:robustness}.) 

It should also be pointed out that the whitening matrix $\mR$ in Eq.~\eqref{eq:ZCA} could be viewed as a $1\times1$ conv functioning on the feature, but directly applying a $1\times1$ conv with learnable parameters on features before computing the standard cost volume cannot replace the proposed learned cost volume. Because $\mR^\top\mR$ only gives a positive semi-definite matrix even when $\mR$ is full-rank, which does not meet the positive definiteness property of an inner product.

% \Uspace
\subsection{Relation with the Weighted Sum of Squared Difference}
% \Lspace

The learnable cost volume can be also formulated by
re-thinking the simplest matching criterion for comparing two features,
\ie the weighted sum of squared difference (WSSD):
\begin{align}\label{eq:ssd}
    \sum_i \lambda_i \left( G_{i}(\mF^2)-G_{i}(\mF^1) \right)^2,
\end{align}
where $G: \R^{c} \rightarrow \R^{c}$ denotes a transformation function on the features
$\mF^{i} \in \R^{c},~i=1,2$, and $G_{i}(\mF)$ indicates the $i^{th}$ element of $G(\mF)$.

By the Taylor series expansion, we have
\begin{align}\label{eq:auto-correlation}
    \sum_i \lambda_i \left( G_{i}(\mF^2)-G_{i}(\mF^1) \right)^2
    \approx \sum_i \lambda_i \left( \nabla G_{i}(\mF^1)^\top \Delta \mF \right)^2
    = \Delta \mF^\top \mW \Delta \mF,
\end{align}
where $\Delta \mF {=} \mF^{2} - \mF^{1}$ is the feature difference and
$\mW = \sum_{i} \lambda_i \nabla G_{i}(\mF^1) \nabla G_{i}(\mF^1)^\top$
is the auto-correlation matrix.
Here, $\mW$ coincides with the kernel matrix of the proposed LCV module in
Eq.~\eqref{eq:learnable_correlation_volume}.
When $\lambda_i=1 (i=1,\ldots,c)$ and $G$ is an identity map,
then $\mW=\mI$, which corresponds to the vanilla cost volume.
If we further expand Eq.~\eqref{eq:auto-correlation}, we can see the connection
with the proposed learnable correlation volume as follows:
\begin{align}\label{eq:connection}
\begin{split}
    \Delta \mF^\top \mW \Delta \mF
    &= (\mF^2-\mF^1)^\top\mW (\mF^2-\mF^1)\\
    &= (\mF^2{}^\top \mW \mF^2 + \mF^1{}^\top \mW \mF^1) - 2\mF^1{}^\top\mW \mF^2,
\end{split}
\end{align}
where the last term shares the same formula with the proposed learnable cost volume.
This implies that the proposed learnable cost volume is inversely correlated with  WSSD. 
As WSSD measures the discrepancy between two features,
the learnable cost volume characterizes a certain kind of similarity between them.

% \Uspace
\section{Experiments}
% \Lspace

% \ds{Find and show examples where LCV helps most}
In this section, we present the experimental results of optical flow estimation
in both supervised and unsupervised settings to demonstrate the effectiveness
of the proposed learnable cost volume.
Also, we carry out ablation studies to show that the LCV module performs favorably against other counterparts.
Moreover, we analyze the behavior of LCV and find it beneficial to
% \yjw{``find it beneficial to" to ``show its advantage on"?}
handling three challenging cases.
More results can be found in the supplementary material and
the source code and trained models will be made available to the public.

\begin{table}[!ht]
  \centering
  \caption{Results of the supervised methods on the MPI Sintel and KITTI 2015 optical flow benchmarks.
    All reported numbers indicate the average endpoint error (AEPE) except for the last two columns,
    where the percentage of outliers averaged over all groundtruth pixels (Fl-all) are presented.
    ``-ft" means finetuning on the relative MPI Sintel or KITTI training set and
    the numbers in the parenthesis are results that train and test on the same dataset.
    Missing entries (-) indicate that the results are not reported for the respective method.
    The best result for each metric is printed in bold.
    }
    \begin{tabular}{l|cc|cc|c|cc}
    \toprule
    \multicolumn{1}{c|}{\multirow{3}[1]{*}{Methods}} & \multicolumn{4}{c|}{Sintel}  & \multicolumn{3}{c}{KITTI 2015} \\
    \cline{2-8}
     & \multicolumn{2}{c|}{Clean} & \multicolumn{2}{c|}{Final} & \multicolumn{1}{c|}{AEPE} &  \multicolumn{2}{c}{Fl-all (\%)} \\ 
    \cline{2-8}
     & train & test & train & test & train & train & test \\
    \hline
    FlowNet2~\cite{ilg2017flownet} & 2.02 & 3.96 & 3.14 & 6.02 & 10.06 &  30.37 & - \\
    FlowNet2-ft~\cite{ilg2017flownet} & (1.45) & 4.16 & (2.01) & 5.74 & (2.30) &  (8.61) & 10.41  \\
    DCFlow~\cite{XRK2017} & - & 3.54 & - & 5.12 & - & 15.09 & 14.83 \\
    MirrorFlow~\cite{Hur2017MirrorFlowES} & - & - & - & 6.07 & - & 9.93 & 10.29 \\
    SpyNet~\cite{ranjan2017optical} &  4.12 & 6.69 & 5.57 & 8.43  & - &  - & - \\
    SpyNet-ft~\cite{ranjan2017optical} &  (3.17) &  6.64 & (4.32) & 8.36 & - & - & 35.07 \\
    LiteFlowNet~\cite{hui2018liteflownet}  & 2.52  & - & 4.05 & 10.39 & -  & - & - \\
    LiteFlowNet+ft~\cite{hui2018liteflownet} & (1.64) & 4.86 & (2.23) & 6.09 &  (2.16) & - &  10.24  \\
    PWC-Net~\cite{sun2018pwc} & 2.55 & - & 3.93 & - &  10.35 & 33.67 & - \\
    PWC-Net-ft~\cite{sun2018pwc} & (2.02) & 4.39 & (2.08) & 5.04 & (2.16) & (9.80) & 9.60 \\
    PWC-Net+-ft~\cite{Sun2019ModelsMS} & (1.71) & 3.45 & (2.34) & 4.60 & (1.50) & (5.30) & 7.72 \\
    IRR-PWC-ft~\cite{hur2019iterative} & (1.92) & 3.84 & (2.51) &  4.58 & (1.63) &  (5.30) & 7.65\\
    HD$^3$~\cite{Yin_2019_CVPR} & 3.84 & - & 8.77 & - & 13.17 & 23.99 & -\\
    HD$^3$-ft~\cite{Yin_2019_CVPR} & (1.70) & 4.79 & (1.17) & 4.67 & (1.31) & (4.10) & 6.55 \\
    VCN~\cite{yang2019vcn} & 2.21 & - & 3.62 & - & 8.36 & 25.10 & 8.73 \\
    VCN-ft~\cite{yang2019vcn} & (1.66) & 2.81 & (2.24) & 4.40 & (1.16) & (4.10) & 6.30 \\
    RAFT~\cite{teed2020raft} & 1.09 & 2.77 & 1.53 & 3.61 & (1.07) & (3.92) & 6.30 \\
    RAFT (warm start)~\cite{teed2020raft} & 1.10 & {\bf 2.42} & 1.61 & 3.39 & - & - & - \\
    \hline
    VCN+LCV & (1.62) & 2.83 & (2.22) & 4.20 & (1.13) & (3.80) & {\bf 6.25} \\
    RAFT+LCV & {\bf(0.94)} & 2.75 & {\bf(1.31)} & 3.55 & {\bf(1.06)} & {\bf(3.77)} & 6.26 \\
    RAFT+LCV (warm start) & (0.99) & 2.49 & (1.47) & {\bf 3.37} &  - & - & -  \\
    \bottomrule
  \end{tabular}%
  \label{tab:flow-supervised}%
%   \vspace{-12pt}
\end{table}

% \PARAspace
\subsubsection{Training Process.}
It is well-known that the deep optical flow estimation pipeline consists the following
stages in the supervised settings~\cite{Sun2019ModelsMS}:
1) train the model on the FlyingChairs~\cite{DFIB15} dataset;
2) finetune the model on the FlyingThings3D~\cite{MIFDB16} dataset; and
3) finetune the model on the Sintel~\cite{Butler:ECCV:2012} and
KITTI~\cite{Menze2018JPRS,Menze2015ISA} training sets.
Besides, there are lots of tricks such as data augmentation and learning rate disruption,
making the training process more complicated.

To avoid the tedious training procedure over multiple datasets,
we adopt a more efficient way to train the model equipped with LCV.
As mentioned in Sec.~\ref{sec:interpretation}, the vanilla cost volume is a special case
of the learnable cost volume when $\mW=\mI$,
which means that the learnable cost volume is more general and
backward compatible with vanilla cost volume.
Therefore, we initialize the kernel matrix $\mW$ as the identity matrix
and other parameters are directly restored from the pre-trained models without using LCV.
After that, we finetune the model with LCV on the Sintel or KITTI datasets
using the same loss function.
This training process not only significantly reduces training time
but also plays a crucial role in the success under the unsupervised settings.
(See Sec.~\ref{sec:unsupervised}.)
This approach can also be viewed as fixing the kernal matrix as $\mW=\mI$
in the first three training stages, and let $\mW$ be learnable in the final stage.

\begin{figure}[!htb]
\centering
\def\picwidth{0.35\textwidth}

\begin{tabular}{cc@{\hskip 0.02\textwidth}c}
\makecell*[c]{Inputs\\\hline AEPE} & 
\makecell*[c]{\includegraphics[width=\picwidth]{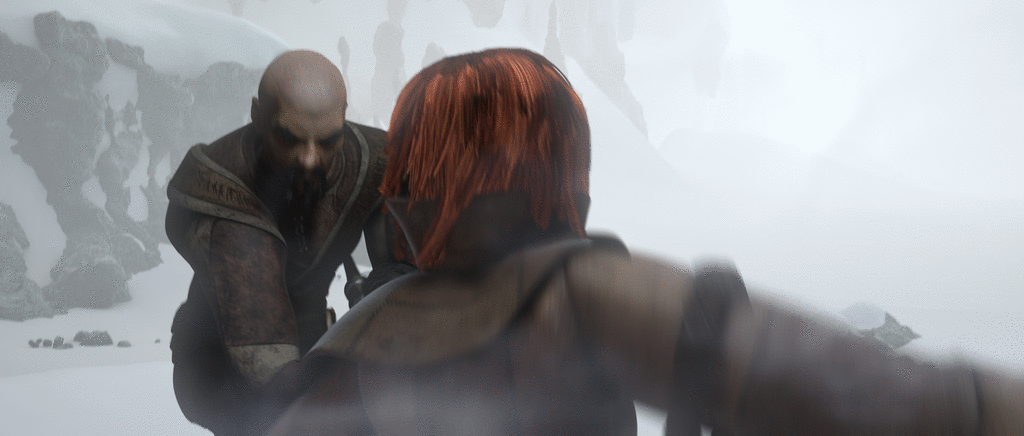}} &
\makecell*[c]{\includegraphics[width=\picwidth]{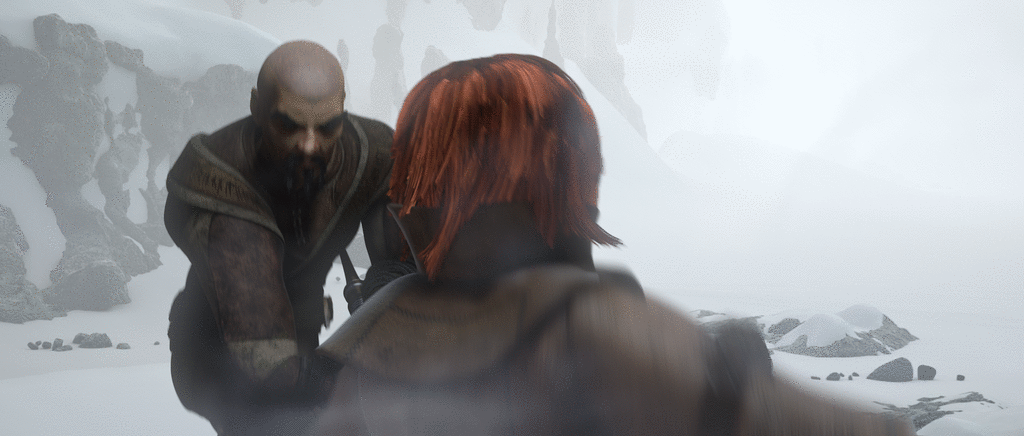}} \\

\makecell*[c]{PWC-Net\\\hline 36.322} & 
\makecell*[c]{\includegraphics[width=\picwidth]{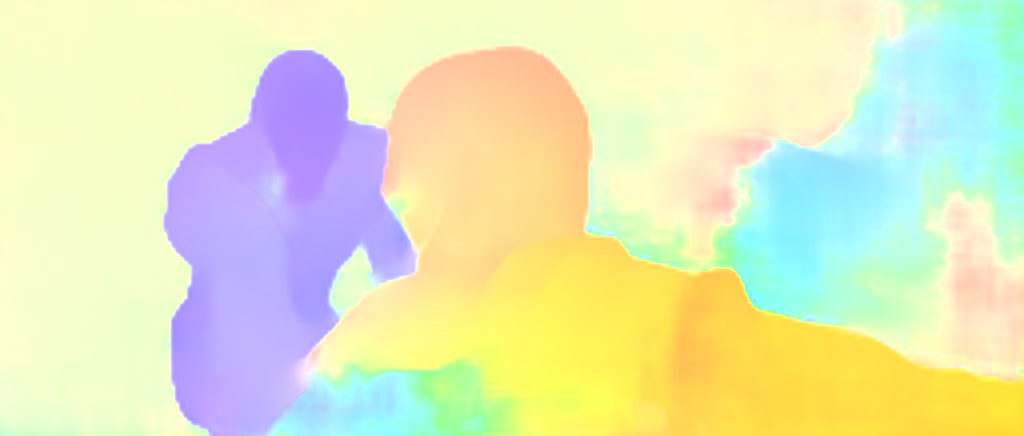}} & 
\makecell*[c]{\includegraphics[width=\picwidth]{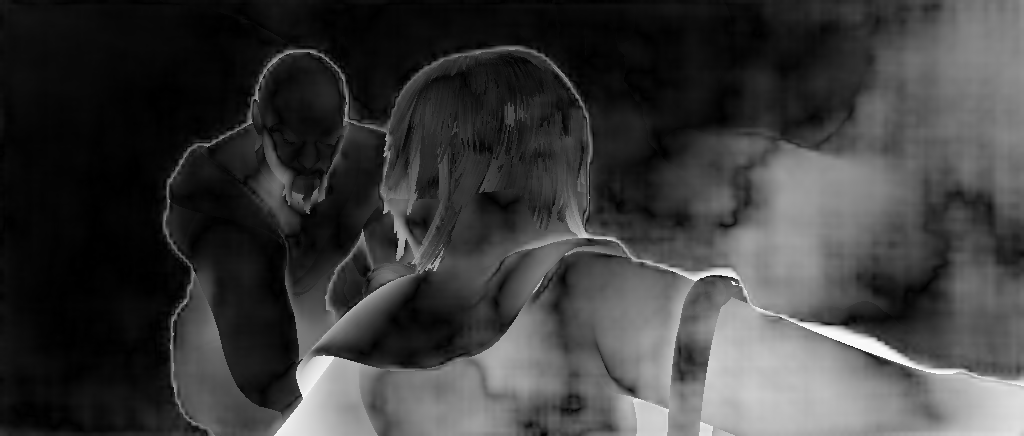}} \\

\makecell*[c]{HD$^3$\\\hline 35.499} & 
\makecell*[c]{\includegraphics[width=\picwidth]{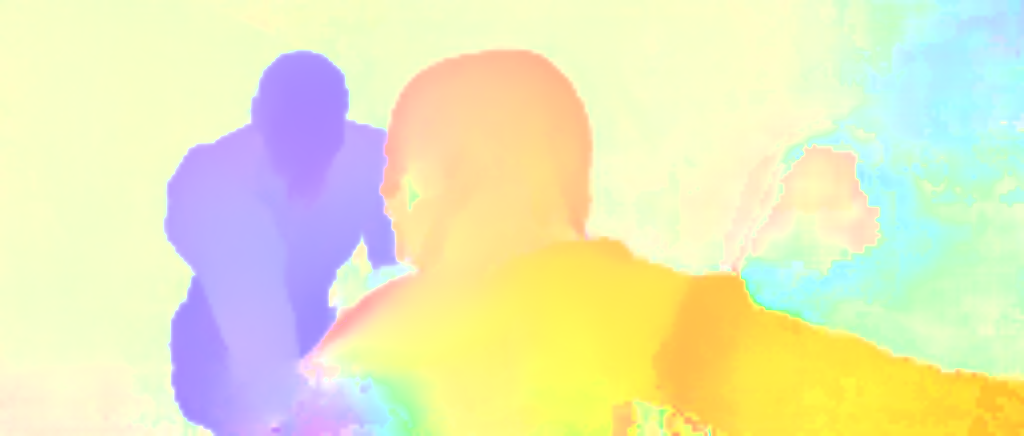}} &
\makecell*[c]{\includegraphics[width=\picwidth]{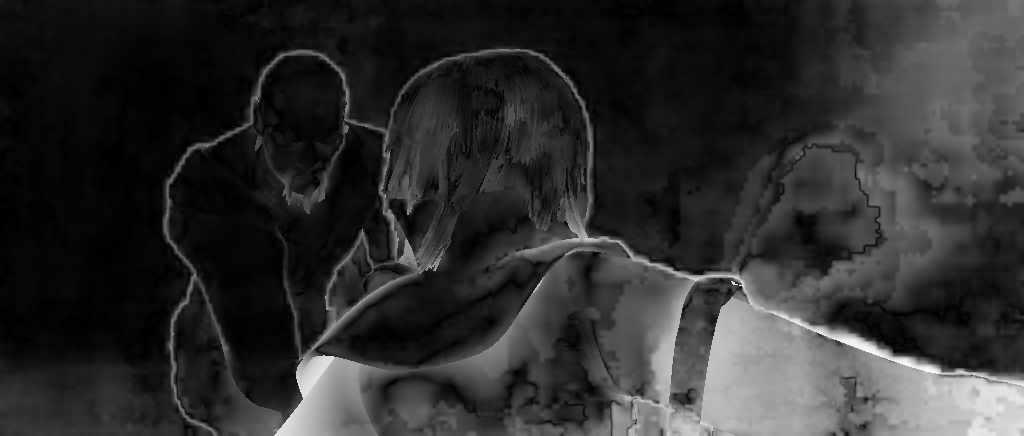}} \\

\makecell*[c]{VCN\\\hline 32.545} & 
\makecell*[c]{\includegraphics[width=\picwidth]{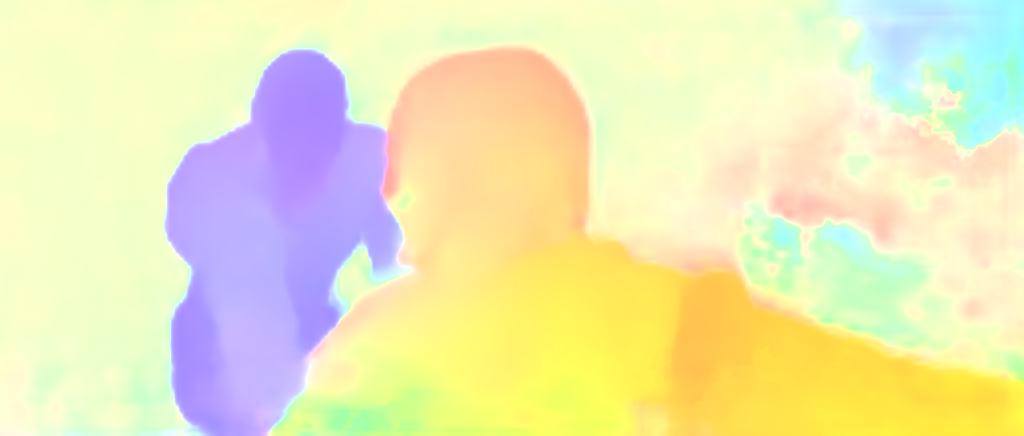}} &
\makecell*[c]{\includegraphics[width=\picwidth]{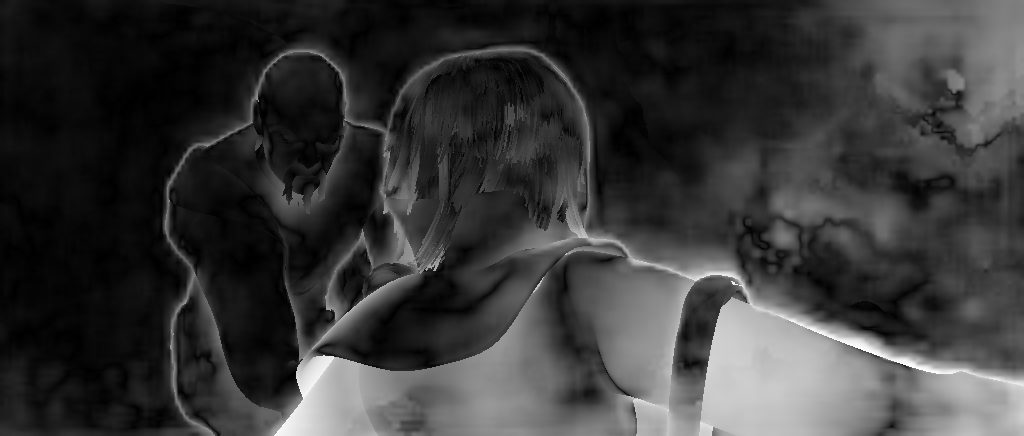}} \\

\makecell*[c]{VCN+LCV\\\hline 28.257} &
\makecell*[c]{\includegraphics[width=\picwidth]{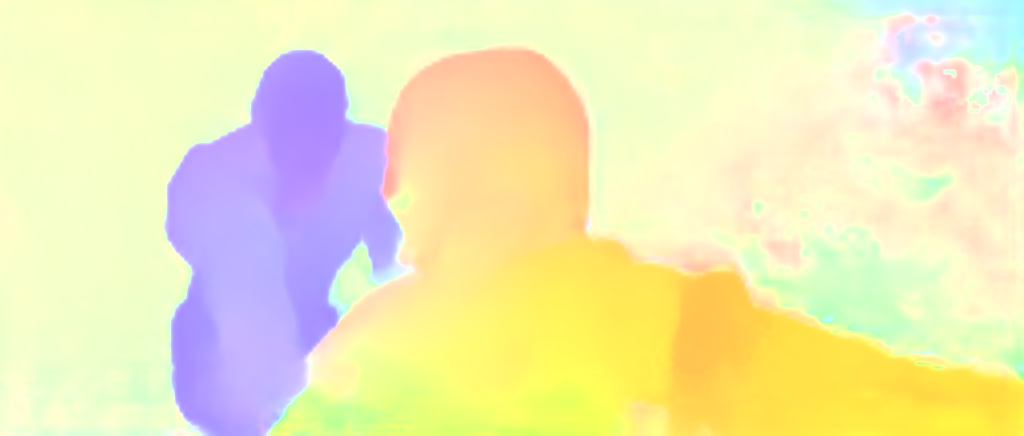}} &
\makecell*[c]{\includegraphics[width=\picwidth]{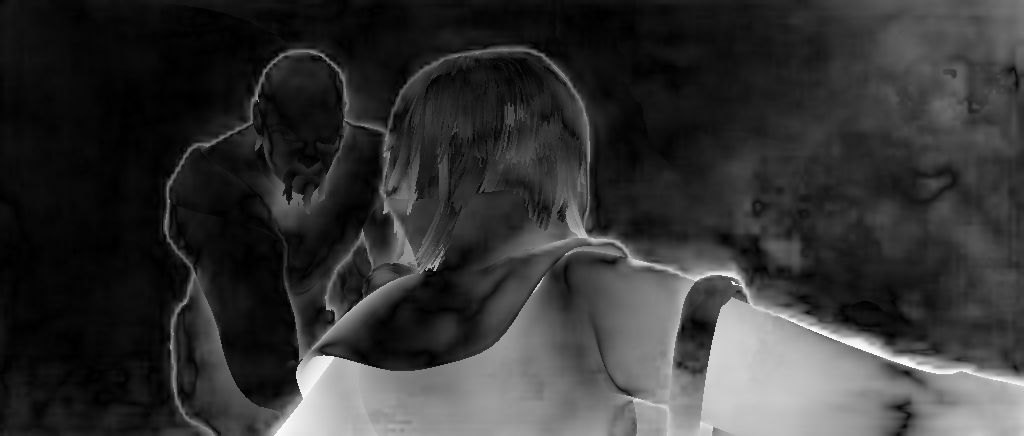}} 
\end{tabular}
\caption{
    Visual results on ``Ambush 1'' from the Sintel test final pass.
    The number under each method denotes the average end-point error (AEPE).
    Left: estimated flow; right: error map (increases from black to white).
}
% \vspace{-12pt}
\label{fig:visualization-sintel}
\end{figure}

\begin{figure}[!htb]
\centering
\def\picwidth{0.35\textwidth}
\begin{tabular}{cc@{\hskip 0.02\textwidth}c}
\makecell*[c]{Inputs\\\hline Fl-all(\%)} &
\makecell*[c]{\includegraphics[width=\picwidth]{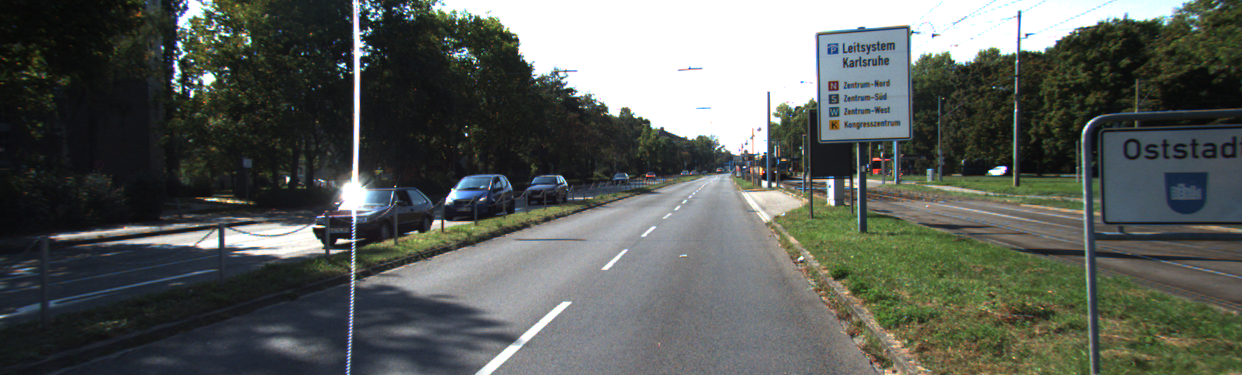}} &
\makecell*[c]{\includegraphics[width=\picwidth]{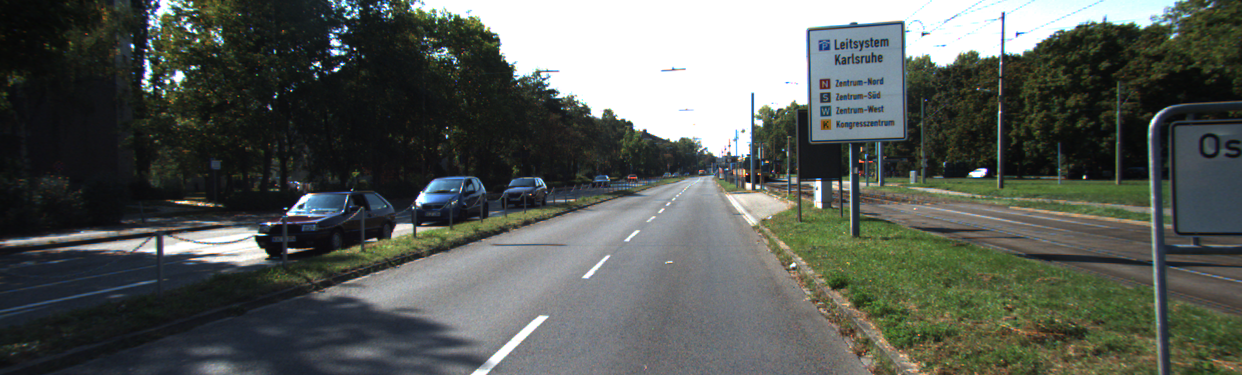}} \\

\makecell*[c]{PWC-Net\\\hline 7.99} & 
\makecell*[c]{\includegraphics[width=\picwidth]{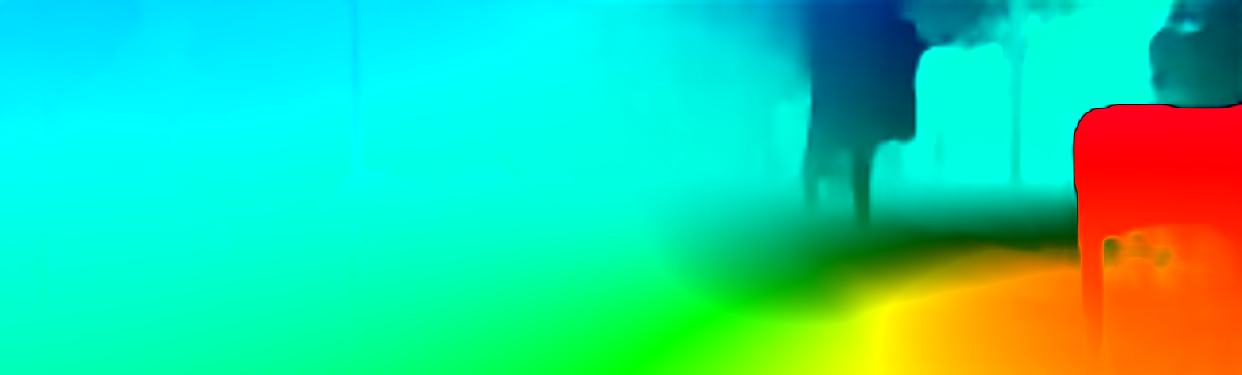}} & 
\makecell*[c]{\includegraphics[width=\picwidth]{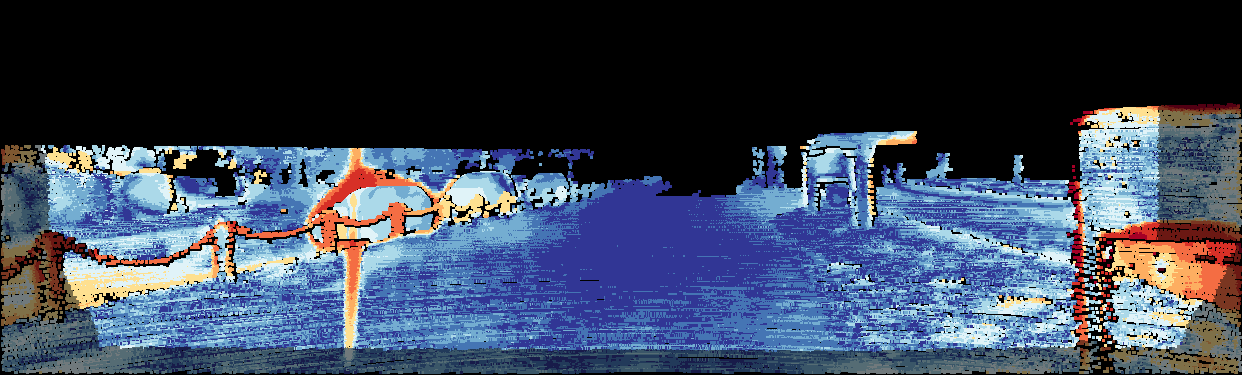}} \\

\makecell*[c]{HD$^3$\\\hline 7.17} & 
\makecell*[c]{\includegraphics[width=\picwidth]{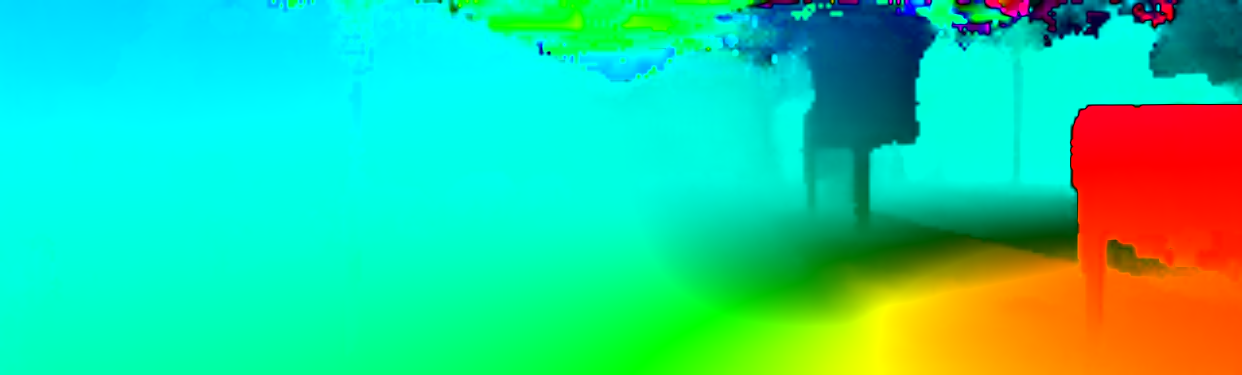}} &
\makecell*[c]{\includegraphics[width=\picwidth]{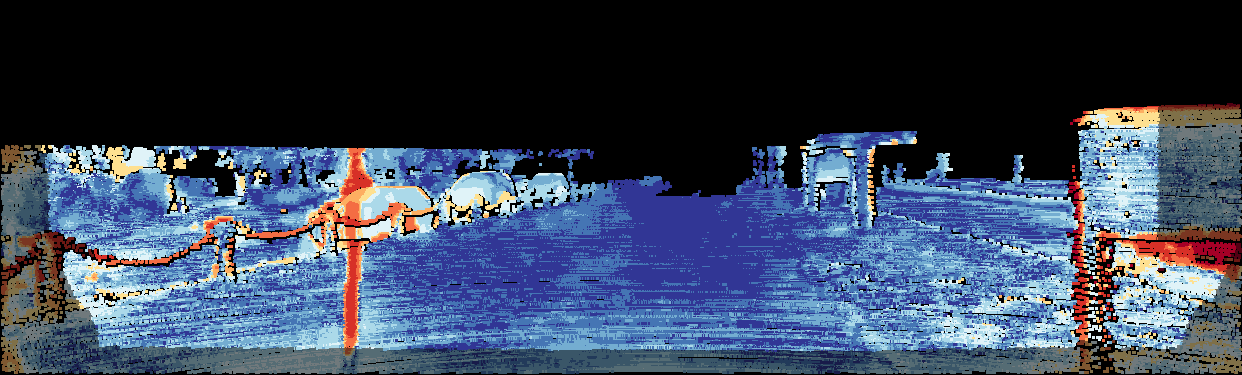}} \\

\makecell*[c]{VCN\\\hline 6.66} & 
\makecell*[c]{\includegraphics[width=\picwidth]{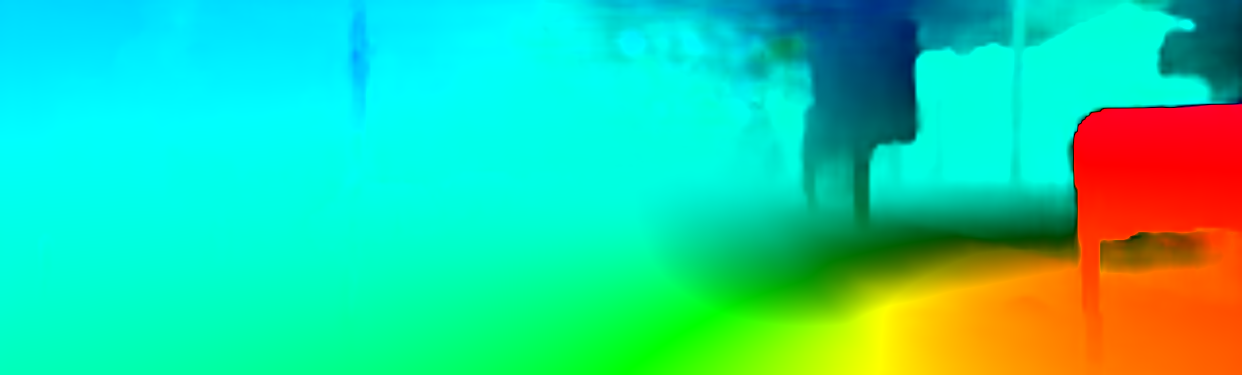}} &
\makecell*[c]{\includegraphics[width=\picwidth]{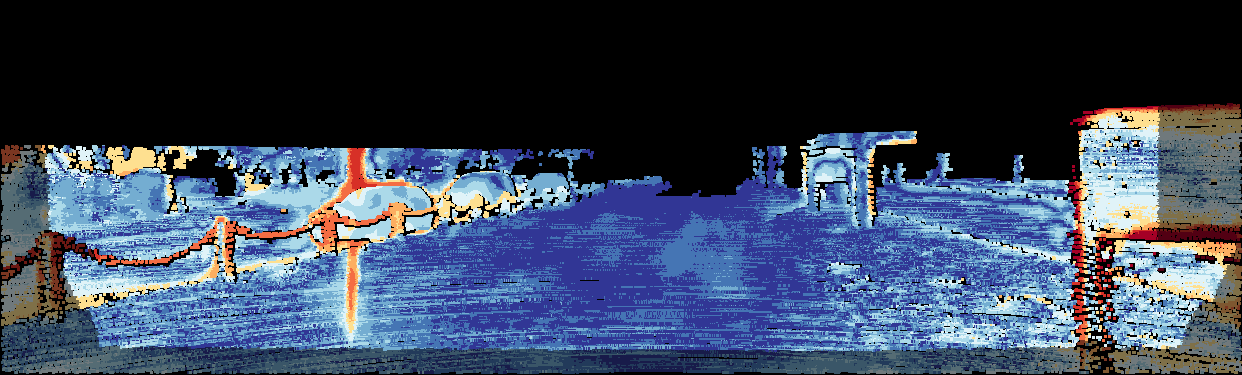}} \\

\makecell*[c]{VCN+LCV\\\hline 6.00} &
\makecell*[c]{\includegraphics[width=\picwidth]{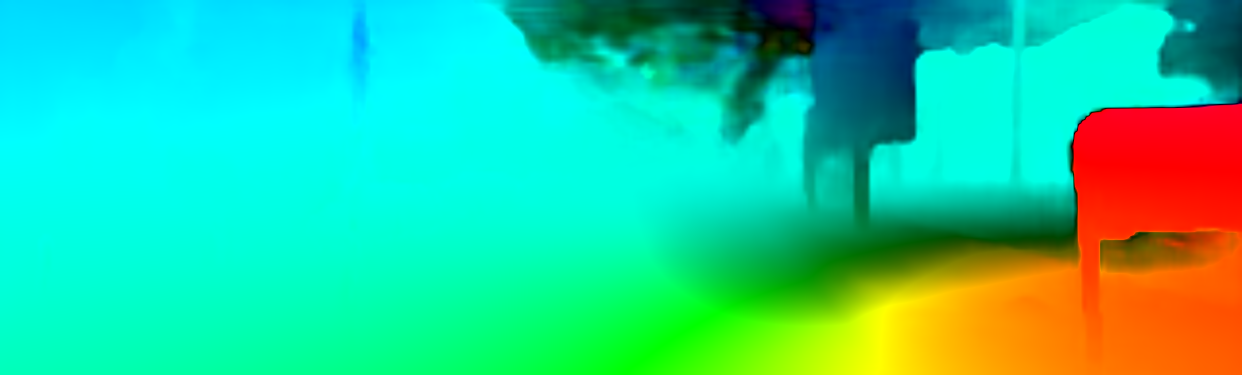}} &
\makecell*[c]{\includegraphics[width=\picwidth]{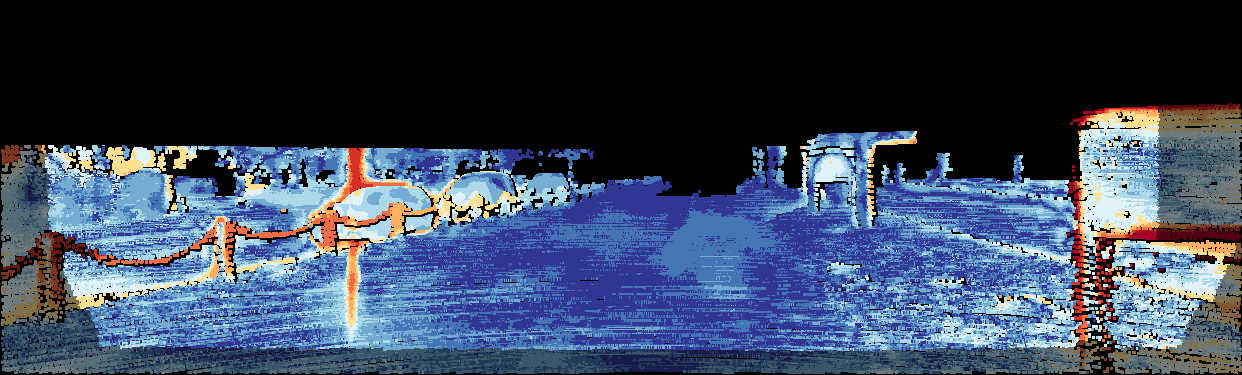}}
\end{tabular}
\caption{
    Visual results on the KITTI 2015 test set.
    The number under each method name denotes the Fl-all score on the given frames.
    Left: estimated flow; right: error map (increases from blue to red).
}
\label{fig:visualization-kitti15}
% \vspace{-12pt}
\end{figure}

% \Uspace
\subsection{Supervised Optical Flow Estimation}
\label{sec:supervised}
% \Lspace

First, we incorporate the learnable cost volume in the VCN~\cite{yang2019vcn} and RAFT~\cite{teed2020raft}
framework, and compare them with other existing methods.
As shown in Table~\ref{tab:flow-supervised}, our method performs favorably
against other state-of-the-art methods on the Sintel Clean/Final pass and the KITTI 2015 benchmark.

The proposed LCV module improves the performance of VCN and RAFT by transforming
the features of video frames to a whitened space to obtain
a clean and robust matching correlation.
This could account for the performance improvement on the Sintel Final pass,
where the scenarios are much harder. 
As shown in Fig.~\ref{fig:visualization-sintel}, 
the flow estimation error for the snow background at the right side
is smaller than other methods.
This is a challenging case because the front person's arm renders
occlusion to part of the snow background and the background
is nearly all white, providing few clues for matching.
However, the LCV module exploits more information from the correlation
among different channels, which assists in obtaining the coherent
flow estimation in the snow background.
The LCV module also has an edge over the vanilla cost volume under the
circumstance of light reflection and occlusion.
As shown in Fig.~\ref{fig:visualization-kitti15}, the prediction error of
our method is smaller around the light reflection region and
the rightmost traffic sign.

Although we do not report the model parameters in the table, the proposed LCV module only makes a very slight increase
in the model size.
The additional parameters come from the kernel matrices
$\mW\in\R^{c\times c}$ at different pyramid levels.
Taking VCN+LCV as an example, there are five kernel matrices in total,
whose channel dimensions are 64, 64, 128, 128, and 128, respectively.
The LCV module only takes up $64^2\times 2+128^2\times 3 = 57,344$ parameters,
which is negligible compared with the entire VCN model of around 6.23M parameters.

% \Uspace
\subsection{Unsupervised Optical Flow Estimation}
\label{sec:unsupervised}
% \Lspace

We also test the LCV module in unsupervised settings
on the KITTI 2015 benchmark.
We replace the vanilla cost volume with the LCV module in the DDFlow~\cite{Liu:2019:DDFlow}
model, and compare it with other unsupervised methods.
As shown in Table~\ref{tab:flow-unsupervised}, our model outperforms the DDFlow baseline,
and even performs favorably against SelFlow~\cite{Liu:2019:SelFlow},
an improved version of DDFlow.

The training process is crucial to the success of the LCV module
in the unsupervised methods. 
Different from the supervised training of optical flow models,
there is no ground truth for direct supervision.
Instead, most unsupervised methods use the photometric loss as a proxy loss.
Specifically, the training of DDFlow  consists of two stages:
1) pre-train a non-occlusion model with census transform~\cite{hafner2013census}, and
2) train an occlusion model by distillation from the non-occlusion model.
If we directly follow the same procedure, the training of DDFlow+LCV
will run into trivial solutions, as the photometric loss does not
give a strong supervision for the correspondence learning,
especially when the LCV module increases the dimension of the solution space.
To prevent from trivial solutions, we fix the kenrel matrix as $\mW=\mI$
in the pre-train stages, and update $\mW$ in the distillation stage.

\begin{table}[t]
  \centering
  \caption{
    Results of the unsupervised methods on the KITTI 2015 optical flow benchmark.
    Missing entries (-) indicate that the results are not reported for the respective method.
    The best result for each metric is printed in bold.
}
    \begin{tabular}{l|c|ccc}
    \toprule
    \multicolumn{1}{c|}{\multirow{3}[1]{*}{Methods}} & \multicolumn{3}{c}{KITTI 2015} \\
    \cline{2-5}
     & \multicolumn{1}{c|}{train} &  \multicolumn{3}{c}{test} \\ 
    \cline{2-5} & \multicolumn{1}{c|}{AEPE} & \multicolumn{1}{c|}{Fl-bg (\%)} & \multicolumn{1}{c|}{Fl-fg (\%)} & \multicolumn{1}{c}{Fl-all (\%)} \\
    \hline
    DSTFlow~\cite{ren2017unsupervised} & 16.79 & - & - & 39 \\
    GeoNet~\cite{Yin2018GeoNetUL} & 10.81 & - & - & - \\
    UnFlow~\cite{meister2018unflow} & 8.88 & - & - & 28.95\\
    DF-Net~\cite{Zou2018DFNetUJ} & 7.45 & - & - & 22.82 \\
    OccAwareFlow~\cite{wang2018occlusion} & 8.88 & - & - & 31.20 \\
    % MultiFrameOccFlow-Hard-ft \cite{janai2018unsupervised} & 6.65 & - & - & - \\
    Back2FutureFlow~\cite{janai2018unsupervised} & 6.59 & 22.67 & 24.27 & 22.94 \\
    SelFlow \cite{Liu:2019:SelFlow} & {\bf 4.84} &  {\bf 12.68} & 21.74 & 14.19 \\
    DDFlow \cite{Liu:2019:DDFlow} & 5.72 & 13.08 & 20.40 & 14.29 \\
    \hline
    DDFlow+LCV (Ours) & 5.15  & 12.98 & {\bf 19.83} & {\bf 14.12} \\
    \bottomrule
    \end{tabular}%
  \label{tab:flow-unsupervised}%

\end{table}

\begin{table}[t]
  \caption{Ablation study of different variants of VCN on the KITTI 2015 dataset.}
  \label{tab:ablation}%
  \def\tabwidth{1\textwidth}
  \centering
  \scalebox{0.8}{
  \begin{tabular}{ccccccccc}
    \toprule
    Methods & VCN & VCN (ct) & VCN ($\mW$, ct) & VCN ($\mLambda$, ct) & VCN ($\mP$, ct) & VCN(1x1 conv) & VCN + LCV \\
    \midrule
    AEPE/Fl-all  & 3.9/1.144 & 4.2/1.204 & 4.1/1.193 & 3.8/1.136 & 3.9/1.129 & 3.9/1.163 & 3.8/1.132 \\
    \bottomrule
  \end{tabular}
  }
%   \vspace{-12pt}
\end{table}

\begin{figure}[!htb]
\centering
\def\picwidth{0.35\textwidth}

\subfloat[Illumination change ($\gamma=0.5$)]{
\begin{tabular}{cc}
\includegraphics[width=\picwidth]{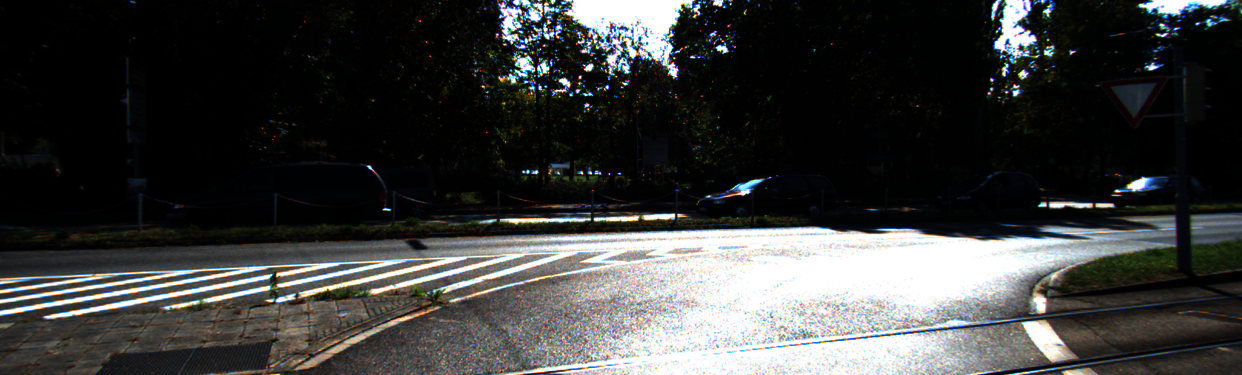} &
\includegraphics[width=\picwidth]{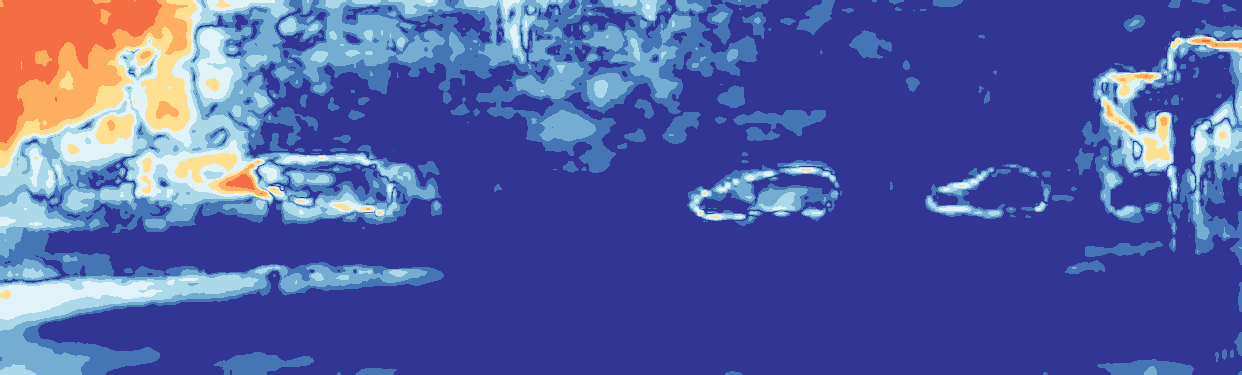} \\
\includegraphics[width=\picwidth]{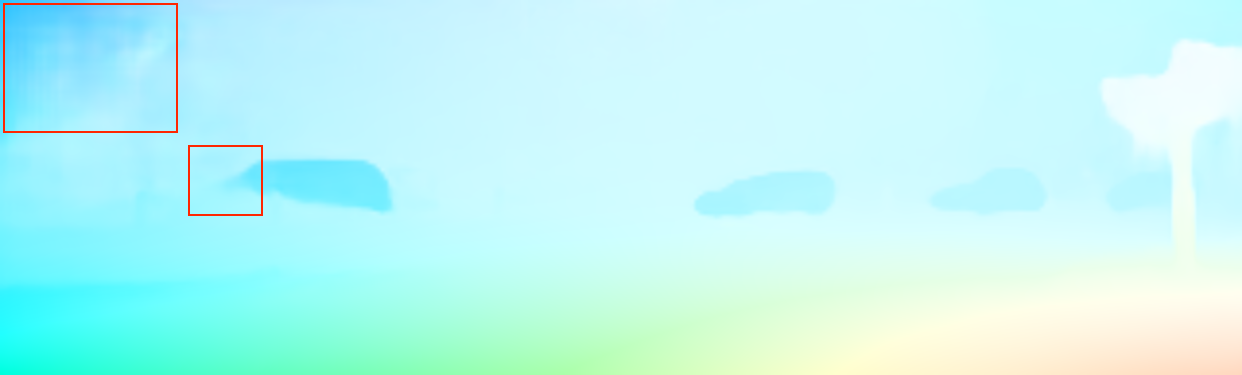} &
\includegraphics[width=\picwidth]{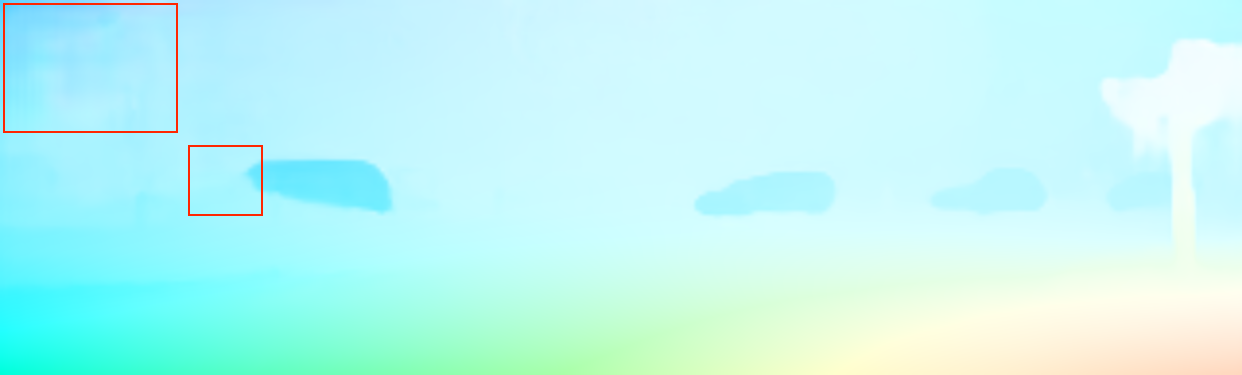} 
\end{tabular}
}

\subfloat[Noise (std=0.001)]{
\begin{tabular}{cc} 
\includegraphics[width=\picwidth]{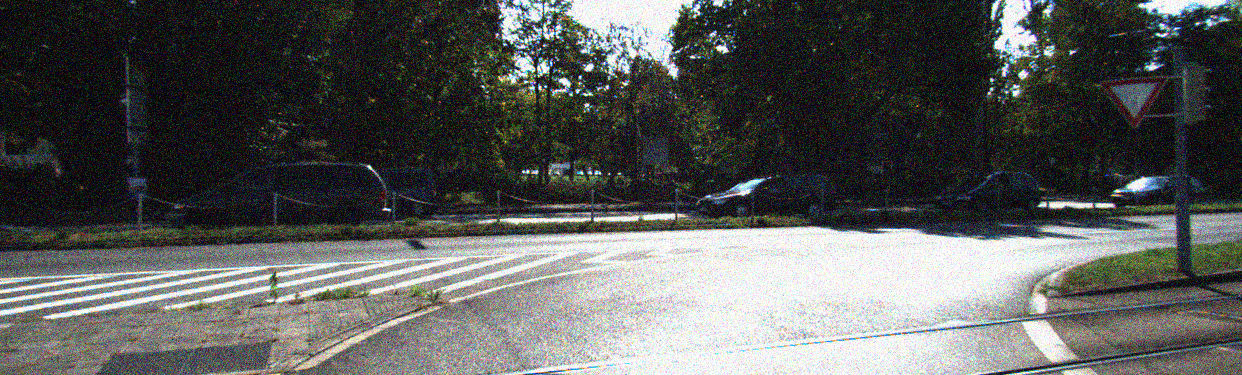} &
\includegraphics[width=\picwidth]{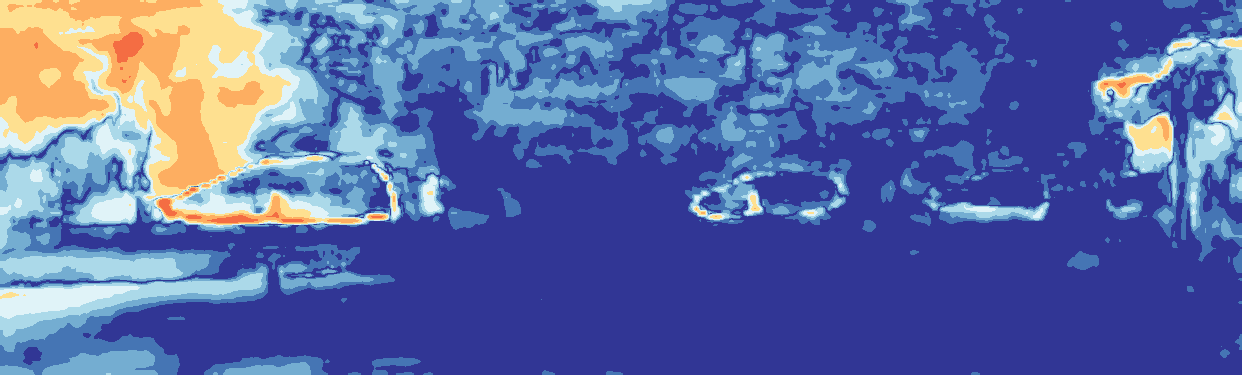} \\
\includegraphics[width=\picwidth]{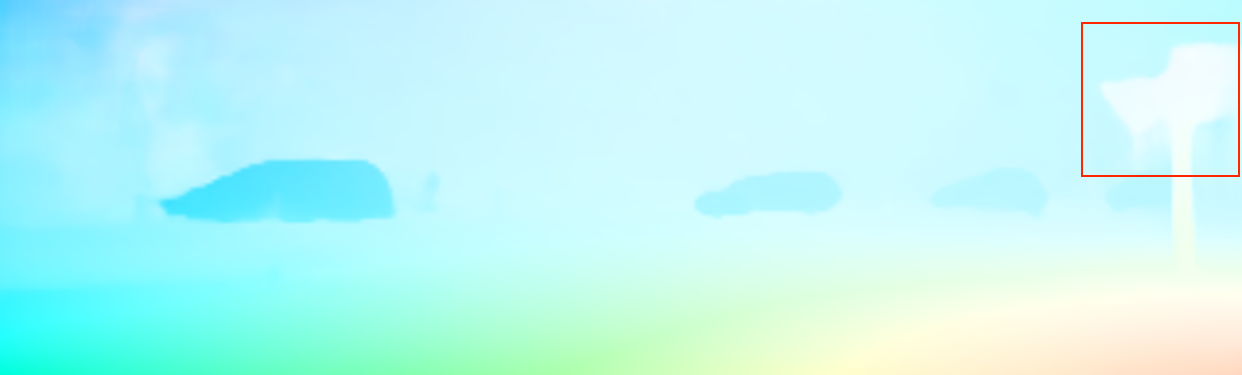} &
\includegraphics[width=\picwidth]{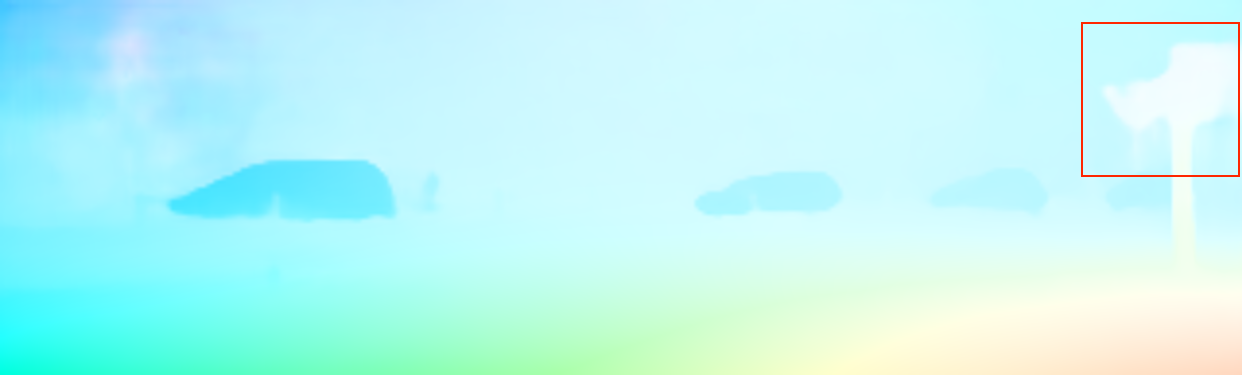} 

\end{tabular}
}

\subfloat[Adversarial patch (radius=50)]{
\begin{tabular}{cc}
\includegraphics[width=\picwidth]{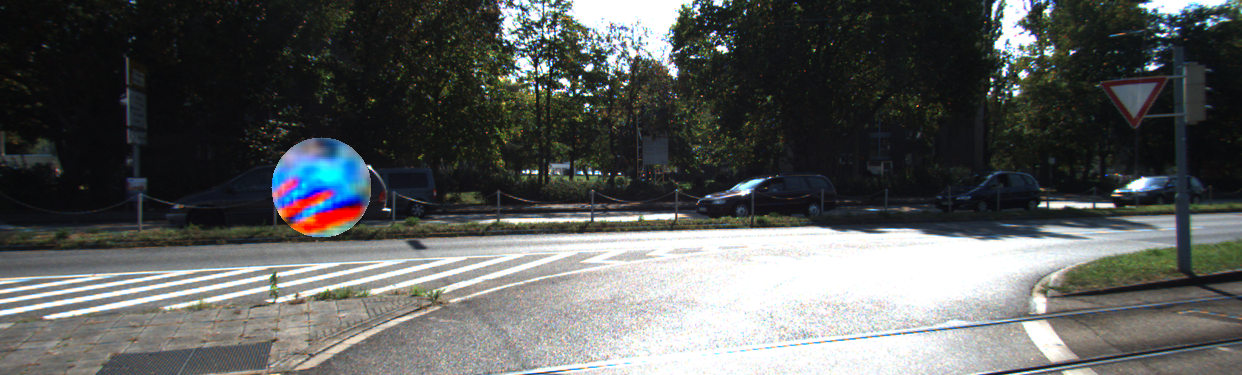} &
\includegraphics[width=\picwidth]{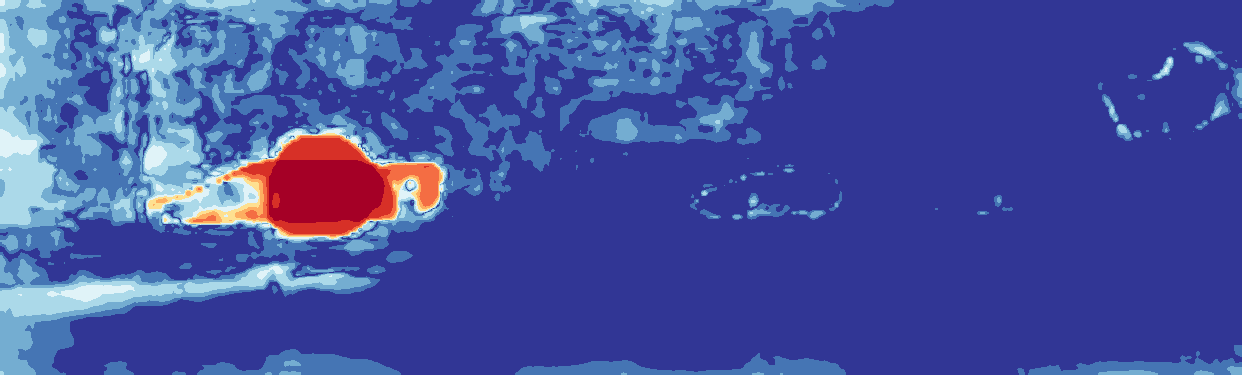} \\
\includegraphics[width=\picwidth]{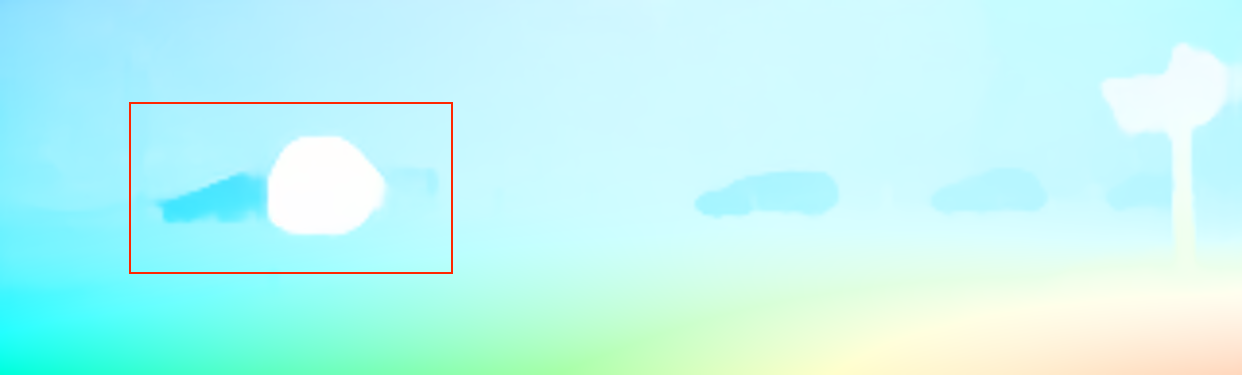} &
\includegraphics[width=\picwidth]{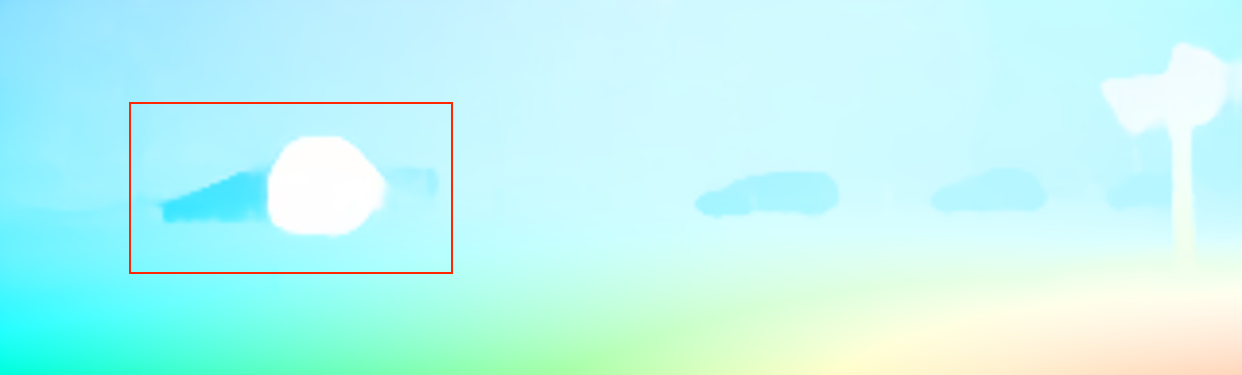} 
\end{tabular}
}
\caption{
    Visual results of three challenging cases, \ie illumination change, noise, and adversarial patch.
    Top left: the first input frame; bottom left/right: flow by VCN / VCN+LCV; 
    top right: flow difference between two methods.
}
\label{fig:visualization-robustness}
% \vspace{-12pt}
\end{figure}

% \Uspace
\subsection{Ablation Study}
\label{sec:ablation}
% \Lspace

We evaluate multiple variants of the LCV module based on the VCN baseline:
\begin{itemize}
    \item VCN: the original VCN baseline.
    \item VCN (ct): continue training the existing VCN using a small learning rate for more epochs.
    \item VCN ($\mW$, ct): remove the symmetry and positive definiteness constraint of $\mW$, \ie, not using the Cayley representation. We restore the weights from the pre-trained VCN and continue training the model with free $\mW$.
    \item VCN ($\mLambda$, ct): fix $\mP$ to be an identity matrix and make the diagonal matrix $\mLambda$ learnable.
    \item VCN ($\mP$, ct): fix $\mLambda$ to be an identity matrix and make the orthogonal matrix $\mP$ learnable.
    \item VCN (1x1 conv): replace the positive definite $\mW$ with $\mR^\top\mR$, where $\mR$ is a $1\times1$ conv operating on features with input and output dimensions equal. $\mR^\top\mR$ is only a positive semi-definite matrix. 
    \item VCN+LCV: employ the Cayley representation to ensure the symmetry and positive definiteness of $\mW$.
\end{itemize}
We randomly split the 200 images with ground truth from the KITTI 2015
training set into the training and validation set by a ratio of 4:1.
As shown in Table~\ref{tab:ablation}, we report the AEPE/Fl-all scores on the validation set.
We observe that continuing training of the VCN model does not bring any benefit,
which indicates that the best VCN model is not obtained at the very end of the training.
Another interesting observation is that VCN ($\mW$, ct) performs better than VCN (ct), showing the benefit of increasing the model capacity.
However, it does not outperform VCN, not even VCN+LCV, confirming the importance of using a valid inner product space.
%
% Although the $\mW$ increases the model capacity,
% whereas the latter part verifies the importance of ensuring a valid inner product space.
%
Comparing the result of VCN (1x1 conv), we can further conclude that ensuring the positive definiteness via the Cayley representation is crucial to the performance. 
We can also find that VCN ($\mLambda$, ct) gets a lower AEPE and VCN ($\mP$, ct) gets a lower Fl-all compared with vanilla VCN. VCN+LCV combines the advantages of both axis rotation and re-weighting, aiming to address two limitations mentioned in the paper.

\begin{table}[t]
  \caption{
    Results on three challenging cases (numbers: AEPE/Fl-all scores).
    %
    % All the numbers in the table denote AEPE/Fl-all scores.
  }
  \label{tab:robustness}%
  \centering
  \subfloat[Illumination change]{
    \scalebox{0.86}{
    \begin{tabular}{c|c|c|c|c|c|c|c|c}
    \toprule
    $\gamma$ & 0.2 & 0.3 & 0.4 & 0.5 & 0.7 & 1.0 & 2.0 & 3.0 \\
    \hline
    VCN  & 16.8/3.240 & 9.9/1.891 & 5.9/1.306 & 3.8/0.995 & 2.7/0.834 & 2.5/0.805 & 2.6/0.819 & 2.6/0.826 \\
    \hline
    VCN+LCV & 17.1/3.232 & 9.8/1.866 & 5.9/1.273 & 3.7/0.967 & 2.6/0.804 & 2.4/0.775 & 2.4/0.790 & 2.5/0.804  \\
    \bottomrule
    \end{tabular}
    }
  }
  
  \begin{tabular}{cc}
    \centering
      \subfloat[Noise]{
        \scalebox{0.69}{
        \begin{tabular}{c|c|c|c|c}
        \toprule
        Standard deviation & 0.0001 & 0.001 & 0.01 & 0.1 \\
        \hline
        VCN  &  2.6/0.816 & 2.9/0.868 & 5.0/1.157 & 19.6/3.213 \\
        \hline
        VCN+LCV & 2.4/0.785 & 2.7/0.838  & 4.7/1.107 & 18.9/3.043 \\
        \bottomrule
        \end{tabular}
        }
      }
      &  
      \subfloat[Adversarial patch]{
        \scalebox{0.69}{
        \begin{tabular}{c|c|c|c|c}
        \toprule
        Patch size & 50 & 100 & 150 & 200 \\
        \hline
        VCN  & 3.5/0.981 & 5.6/1.419 & 8.5/2.048 & 11.9/2.880 \\
        \hline
        VCN+LCV & 3.4/0.949 & 5.5/1.384 & 8.3/2.004 & 11.6/2.801  \\
        \bottomrule
        \end{tabular}
        }
      }
 \end{tabular}
% \vspace{-15pt}
\end{table}

% \Uspace
\subsection{Robustness Analysis}
\label{sec:robustness}
% \Lspace

To further understand the effect of the LCV module, we evaluate the flow estimation
performance under three challenging cases, \ie 1) illumination changes: 
we adjust the illumination of the input frames by changing the value of $\gamma$,
where $\gamma=1.0$ is the original image, $\gamma<1.0$ is for a darker image,
and $\gamma>1.0$ is for a brighter image.
2) adding noises: we adjust the standard deviation to control the noise magnitude.
and 3) inserting adversarial patches: we borrow the universal adversarial patch~\cite{ranjan2019attacking}
that can perform a black-box attack for all optical flow models,
and insert patches of different sizes to the input frames.

We compare the VCN model and its variant equipped with LCV.
Both two models are trained on the KITTI 2015 training set.
For qualitative comparison, we perform the above three types of processing on 194 images with the flow groundtruth from the KITTI 2012 as our test set.
As shown in Table~\ref{tab:robustness}(a), VCN+LCV consistently outperforms the VCN baseline in all three challenging cases.
For better illustration, we visualize the effect on an image from KITTI 2015 test set as shown in Fig.~\ref{fig:visualization-robustness}.
It can be seen that the LCV module can help stabilize the flow prediction
around the background trees at the top left corner of the frame
under the cases of dark illumination and random noise injection.
In the third example, the outline of the car body near the patch circle
is better preserved by our model.
(See the difference map for details.)

% \Uspace
\section{Conclusions}
% \Lspace

In this work, we introduce a learnable cost volume (LCV) module for optical flow estimation.
The proposed LCV module generalizes the standard Euclidean inner product
into an elliptical inner product with a symmetric and positive definite kernel matrix.
To keep its symmetry and positive definiteness, we use the Cayley representation
to re-parameterize the kernel matrix for end-to-end training.
The proposed LCV is a lightweight module and can be easily plugged into
any existing networks to replace the vanilla cost volume.
Experimental results show that the proposed LCV module improves both the accuracy
and the robustness of state-of-the-art optical flow models.

\subsubsection*{Acknowledgement.} This work is supported in part by NSF CAREER Grant 1149783. We also thank Pengpeng Liu and Jingfeng Wu for kind help.

\clearpage
\bibliographystyle{splncs04}

% \bibliography{0872}

\newpage

\appendix

\section{Proofs of Theorems}
\label{sec:proof}

\subsection{Proof of Theorem 1}

The Theorem of Cayley representation was first given by Cayley in his paper~\cite{cayley1846algebraic}. However, the old paper is not available online. For convenience, we give a simple proof here.

\begin{proof}
First, we validate that the $\mP$ is a orthogonal matrix. The condition that $\mS\in\SO^*(n)$ ensures that $(\mI+\mS)$ is invertible. Since $\mS$ is  skew-symmetric, so $\mS^\top\mS = -\mS^2=\mS\mS^\top$. Hence we have
\begin{align}
    \mP^\top\mP &= (\mI + \mS)^{-\top} (\mI - \mS)^\top (\mI - \mS)(\mI + \mS)^{-1} \\
    &= (\mI + \mS^\top)^{-1} (\mI - \mS^\top) (\mI - \mS)(\mI + \mS)^{-1} \\
    &= (\mI - \mS)^{-1}  (\mI - \mS^\top - \mS - \mS^\top\mS)(\mI + \mS)^{-1}\\
    &= (\mI - \mS)^{-1} (\mI - \mS^\top - \mS - \mS\mS^\top)(\mI + \mS)^{-1} \\
    &= (\mI - \mS)^{-1}  (\mI - \mS) (\mI - \mS^\top)(\mI + \mS)^{-1} \\
    &= (\mI - \mS^\top)(\mI + \mS)^{-1} \\
    &= (\mI + \mS)(\mI + \mS)^{-1} \\
    &= \mI.
\end{align}
Next, we need to show the uniqueness of the Cayley representation.
\begin{alignat}{2}
    \phantom{\iff}& \mP &&= (\mI - \mS)(\mI + \mS)^{-1} \\
    \iff & \mP(\mI + \mS) &&= \mI - \mS\\
    \iff & \mP + \mP\mS &&= \mI - \mS\\
    \iff & \mS + \mP\mS &&= \mI - \mP\\
    \iff & (\mI + \mP)\mS &&= \mI - \mP\\
    \iff & \mS &&= (\mI + \mP)^{-1}(\mI - \mP)
\end{alignat}
Therefore, the skew-symmetric matrix $\mS$ is uniquely represented by $\mP$, which concludes the proof. \qed
\end{proof}

\subsection{Proof of Theorem 2}
Before giving the proof, we would like to recall the definition of connectedness.

\begin{defn}[Connectedness]
A set of matrices $\mathcal{G}$ is said to be connected if for all $\mA$ and $\mB$
in $\mathcal{G}$, there exists a continuous path $\mA(t),~0\le t\le 1$,
lying with $\mA(0)=\mA$ and $\mA(1)=\mB$.
\end{defn}

The above definition of connectedness is actually path connectedness
in topology. 
% 
% The $\SO^*(n)$ is not a matrix Lie group, but here we adopt the same topology in $\R^{n^2}$. 
% 
Now we begin our proof of Theorem~2.
\begin{proof}
Since the identity matrix $\mI\in \SO^*(n)$, it suffices to prove that
for any $\mX\in \SO^*(n)$, there exists a continuous path $\mA(t),~0\le t\le 1$, such that $\mA(0)=\mI$ and $\mA(1)=\mX$. For any $\mX \in \SO^*(n)$, we have its spectral decomposition 
\begin{align}
    \mX = \mP^\top \diag(K_1, \ldots, K_q, 1, \ldots, 1) \mP,
\end{align}
where the $\mP \in \mathrm{O}(n)$ and $0 \le q \le n/2$, and 
\begin{align}
    K_\lambda = \left(
    \begin{array}{rr}
        \cos(\theta_\lambda) & -\sin(\theta_\lambda)  \\
        \sin(\theta_\lambda) & \cos(\theta_\lambda)
    \end{array}
    \right), 
    \quad \theta_\lambda\in[-\pi, \pi), \quad \lambda=1,\ldots,q.
\end{align}
If we put
\begin{align}
    K_\lambda(t) = \left(
    \begin{array}{rr}
        \cos(t\theta_\lambda) & -\sin(t\theta_\lambda)  \\
        \sin(t\theta_\lambda) & \cos(t\theta_\lambda)
    \end{array}
    \right), 
\end{align}
then the path required is 
\begin{align}
    \mA(t) = \mP^\top\diag(K_1(t), \ldots, K_q(t), 1, \ldots, 1) \mP.
\end{align}
\qed
\end{proof}

\section{More Results}
\label{sec:add:results}

\subsection{Ranking on Sintel and KITTI 2015 Benchmark}

The ranking results on the Sintel and KITTI 2015 benchmark can be found at
\url{http://sintel.is.tue.mpg.de/results} and \url{http://www.cvlibs.net/datasets/kitti/eval_scene_flow.php?benchmark=flow}. Here we capture the screenshot of the ranking results by March 8, 2020. 

\newpage
\begin{figure}[!htbp]
    \centering
    \def\picwidth{0.9\textwidth}
    \subfloat[Sintel Final]{\includegraphics[width=\picwidth]{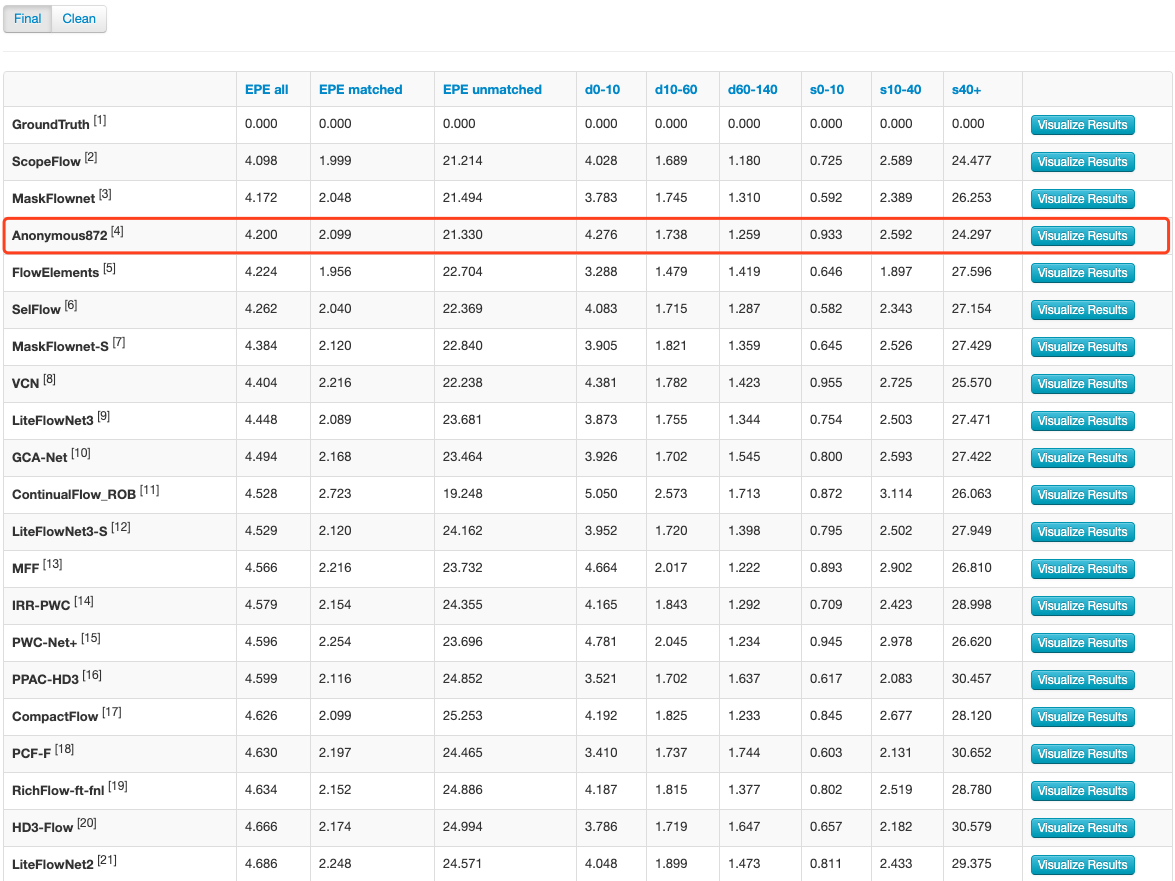}}
    \\
    \subfloat[Sintel Clean]{\includegraphics[width=\picwidth]{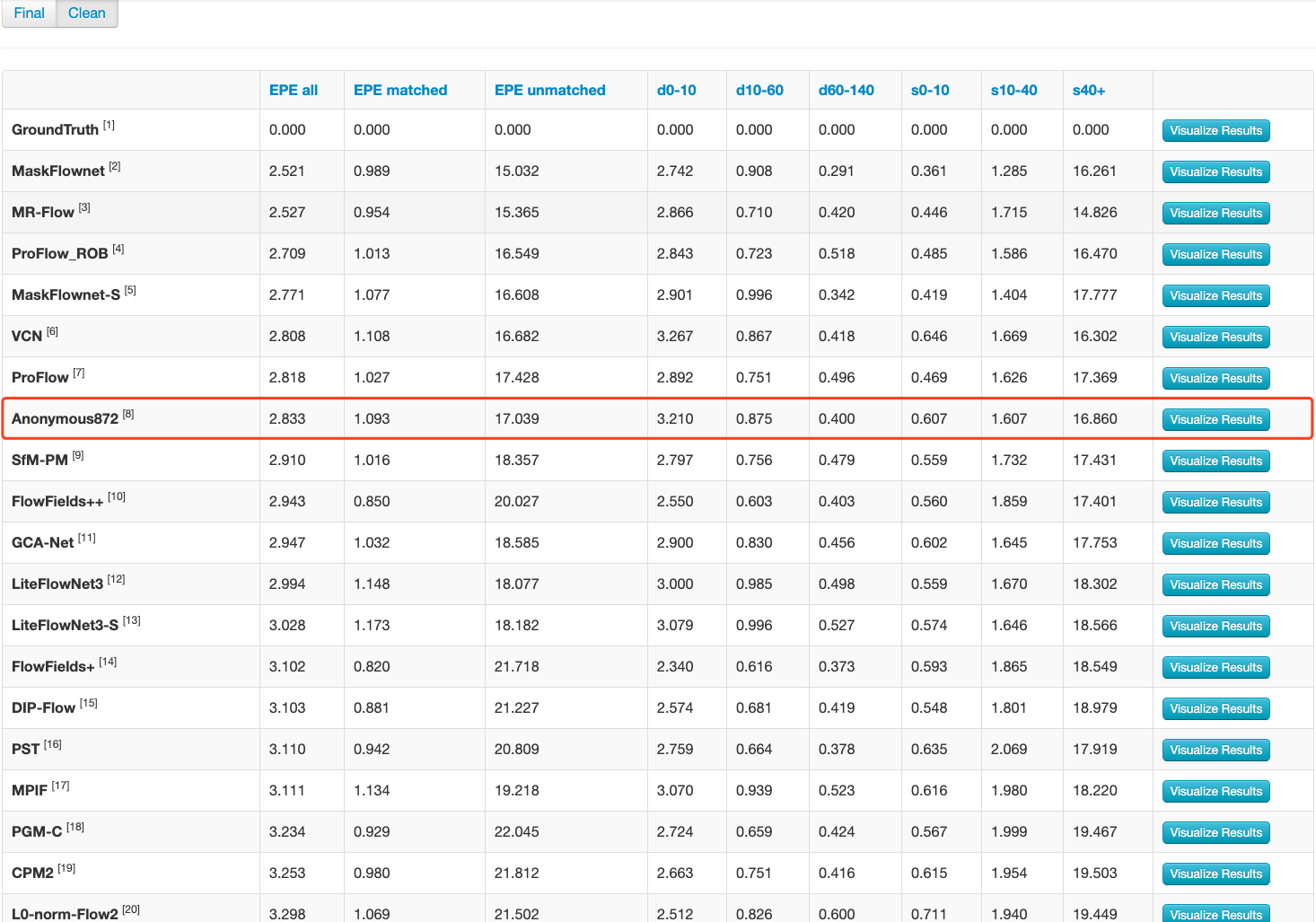}}
    \caption{Ranking results on Sintel and KITTI 2015 benchmark.}
\end{figure}

\begin{figure}[!htbp]
    \ContinuedFloat
    \centering
    \def\picwidth{0.9\textwidth}
    \subfloat[KITTI 2015 - Supervised]{\includegraphics[width=\picwidth]{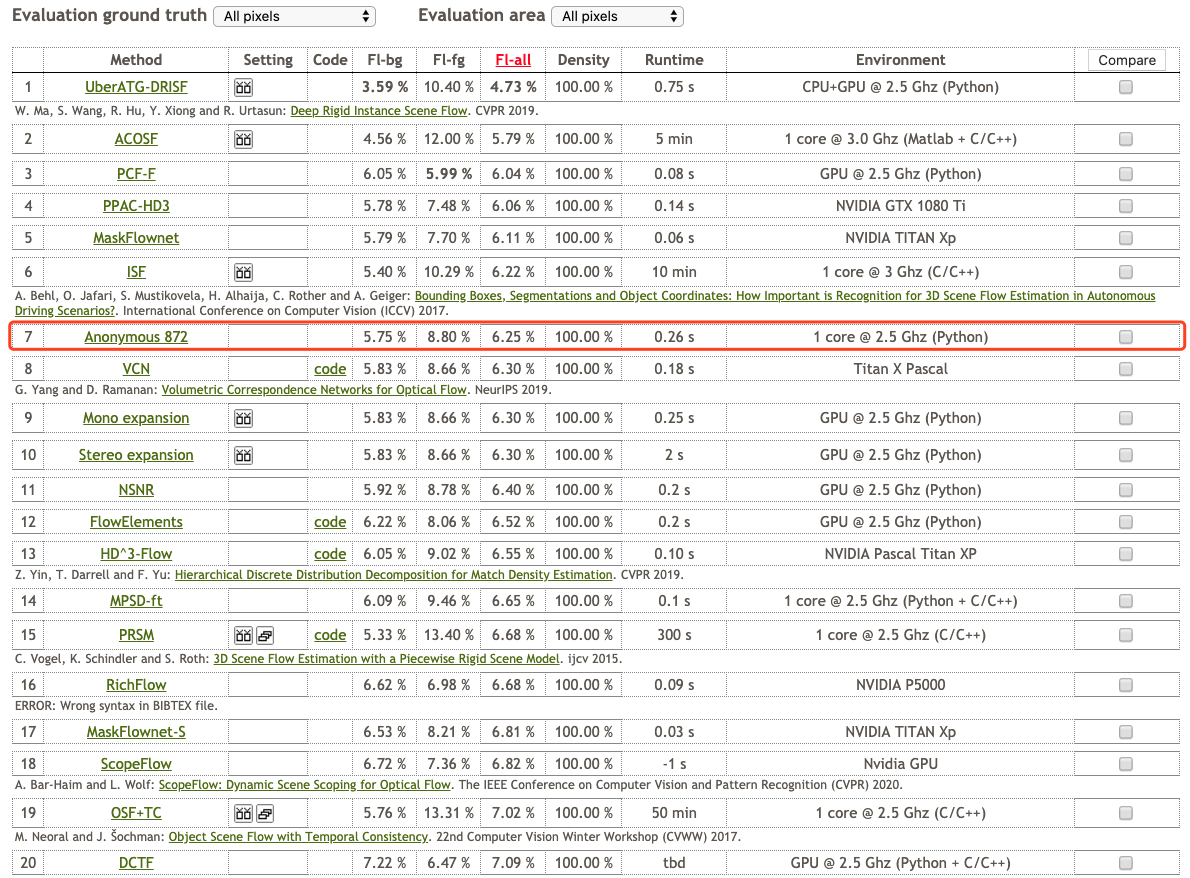}}
    \\
    \subfloat[KITTI 2015 - Unsupervised]{\includegraphics[width=\picwidth]{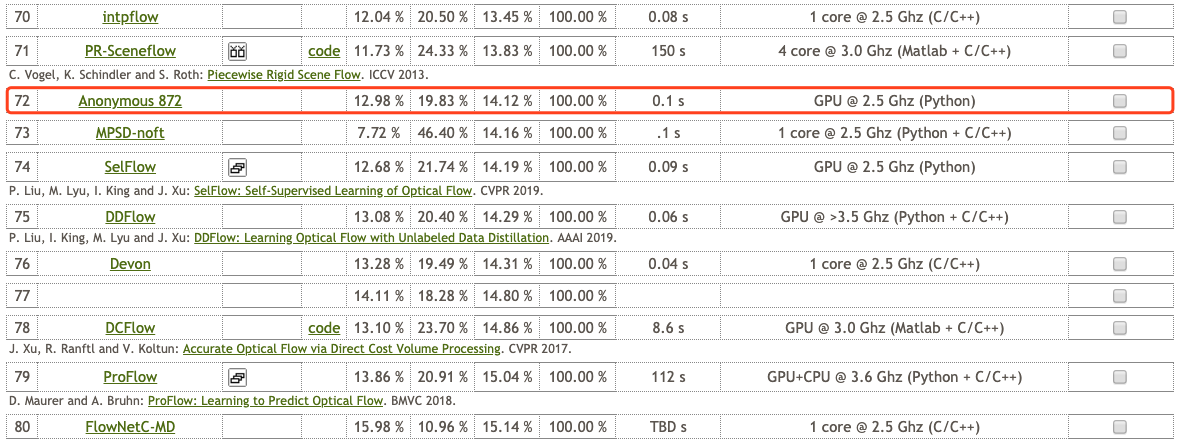}}
    \caption{Ranking results on Sintel and KITTI 2015 benchmark.}
    \label{fig:ranking}
\end{figure}

\subsection{More Visualization Results}

\begin{figure}[!htb]
\centering
\def\picwidth{0.4\textwidth}

\begin{tabular}{cc@{\hskip 0.02\textwidth}c}
\makecell*[c]{Inputs\\\hline AEPE} & 
\makecell*[c]{\includegraphics[width=\picwidth]{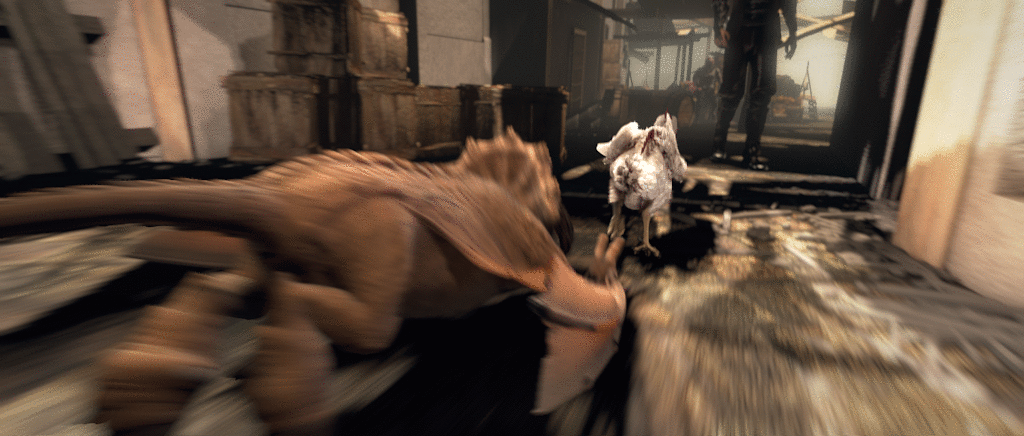}} &
\makecell*[c]{\includegraphics[width=\picwidth]{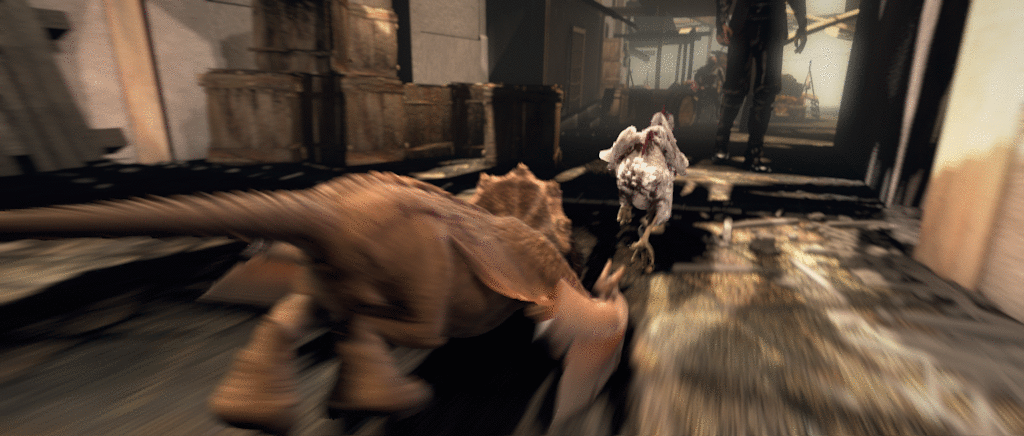}} \\

\makecell*[c]{PWC-Net\\\hline 18.948} & 
\makecell*[c]{\includegraphics[width=\picwidth]{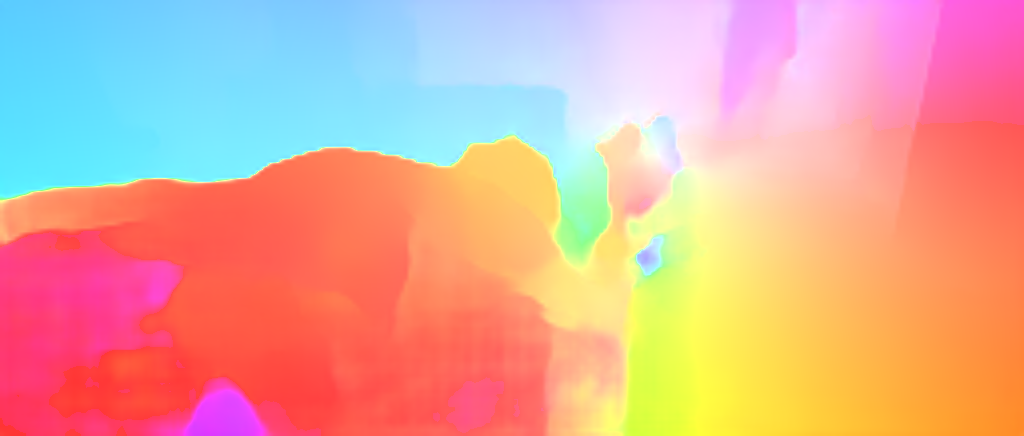}} & 
\makecell*[c]{\includegraphics[width=\picwidth]{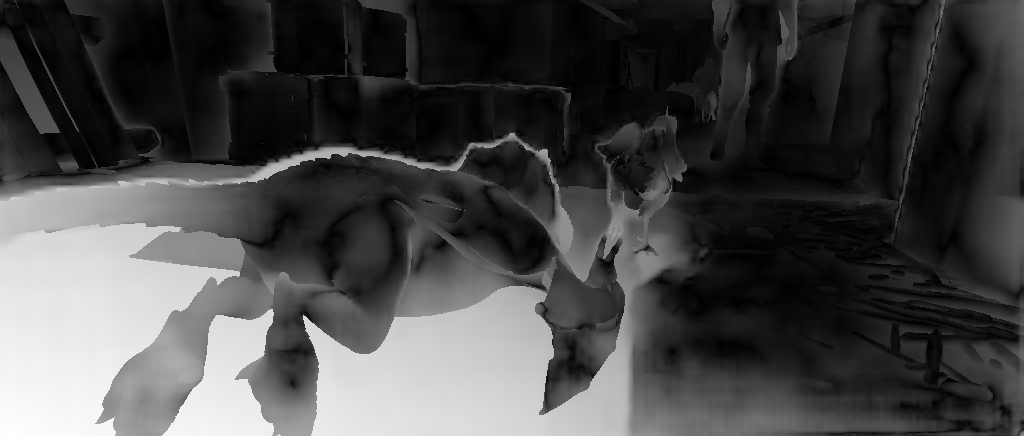}} \\

\makecell*[c]{HD$^3$\\\hline 19.542} & 
\makecell*[c]{\includegraphics[width=\picwidth]{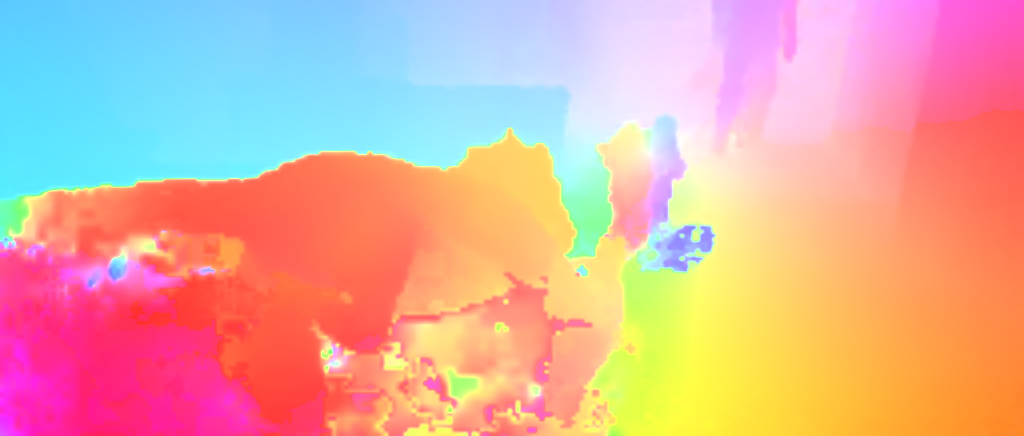}} &
\makecell*[c]{\includegraphics[width=\picwidth]{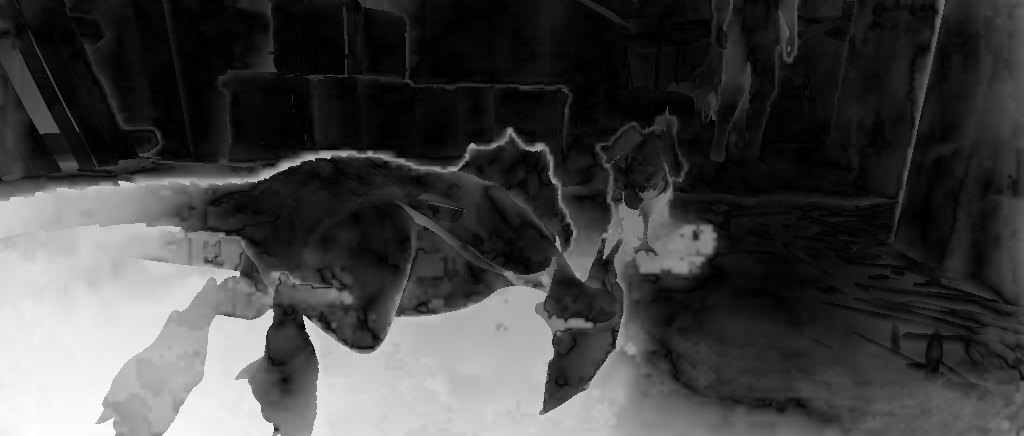}} \\

\makecell*[c]{VCN\\\hline 14.294} & 
\makecell*[c]{\includegraphics[width=\picwidth]{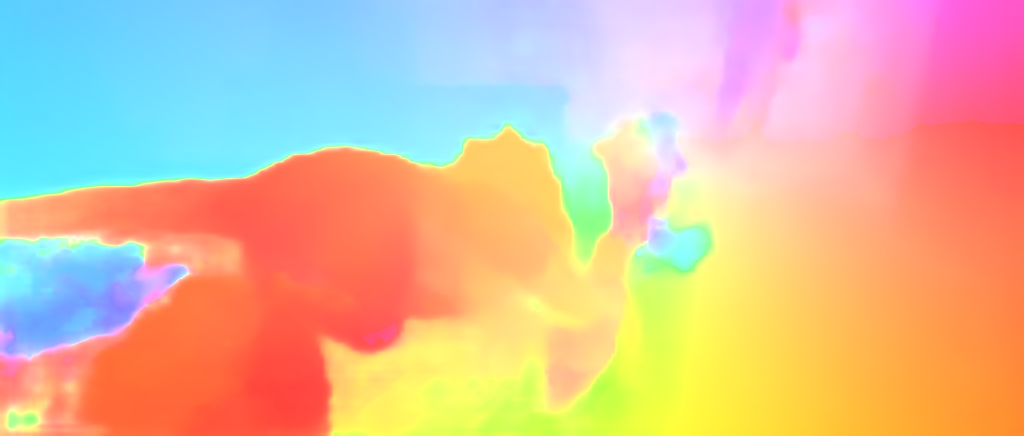}} &
\makecell*[c]{\includegraphics[width=\picwidth]{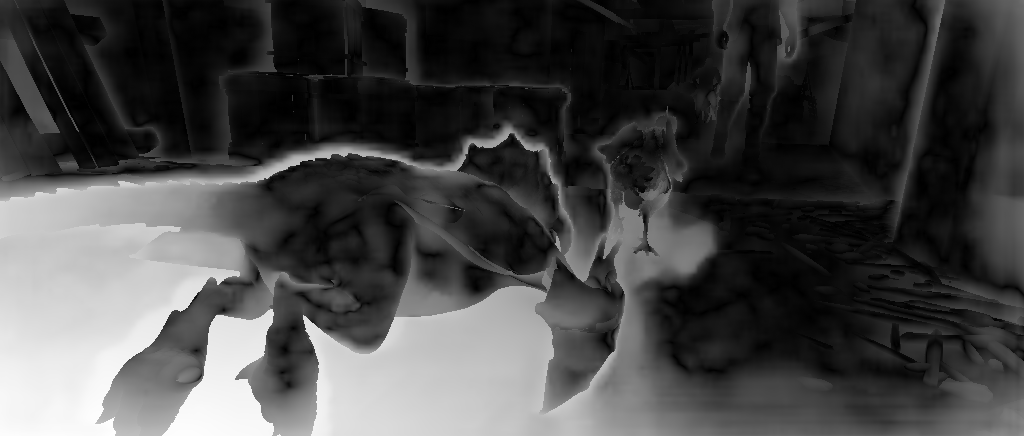}} \\

\makecell*[c]{VCN+LCV\\\hline 14.176} &
\makecell*[c]{\includegraphics[width=\picwidth]{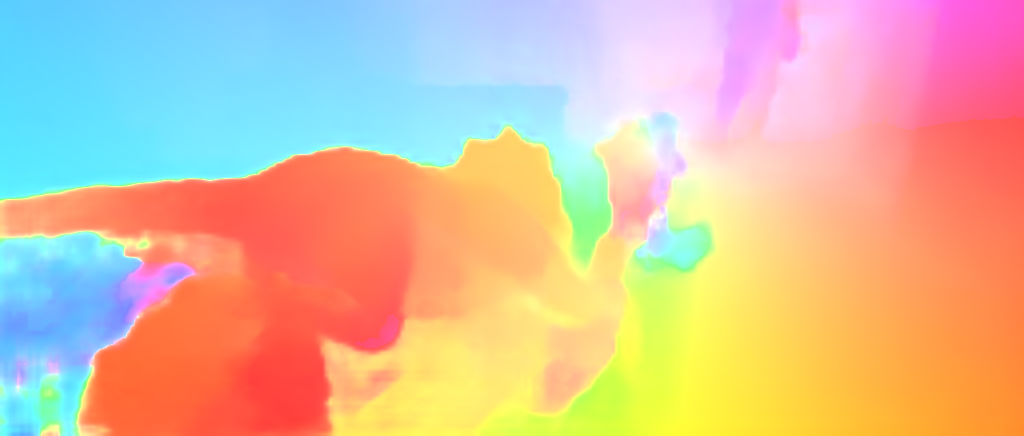}} &
\makecell*[c]{\includegraphics[width=\picwidth]{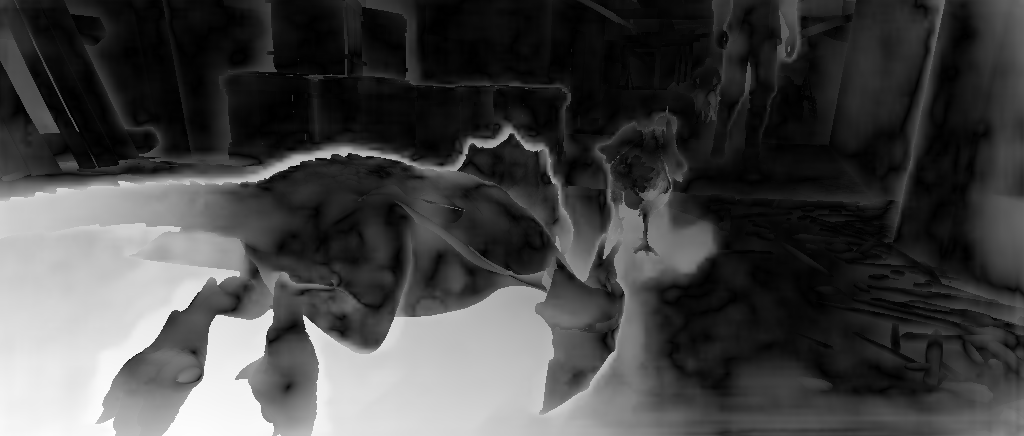}} 
\end{tabular}

\caption{
    More visualization results on ``Market 4'' from the Sintel test final pass.
    The number under each method name denotes the average end-point error (AEPE) on the given frames.
    The estimated flow and error maps are presented on the left and right sides, respectively.
    In the error map, the error of the estimated flow increases from black to white. 
}
\label{fig:more-visualization-sintel}
\end{figure}

\begin{figure}[!htb]
\centering
\def\picwidth{0.4\textwidth}
\begin{tabular}{cc@{\hskip 0.02\textwidth}c}
\makecell*[c]{Inputs\\\hline Fl-all(\%)} &
\makecell*[c]{\includegraphics[width=\picwidth]{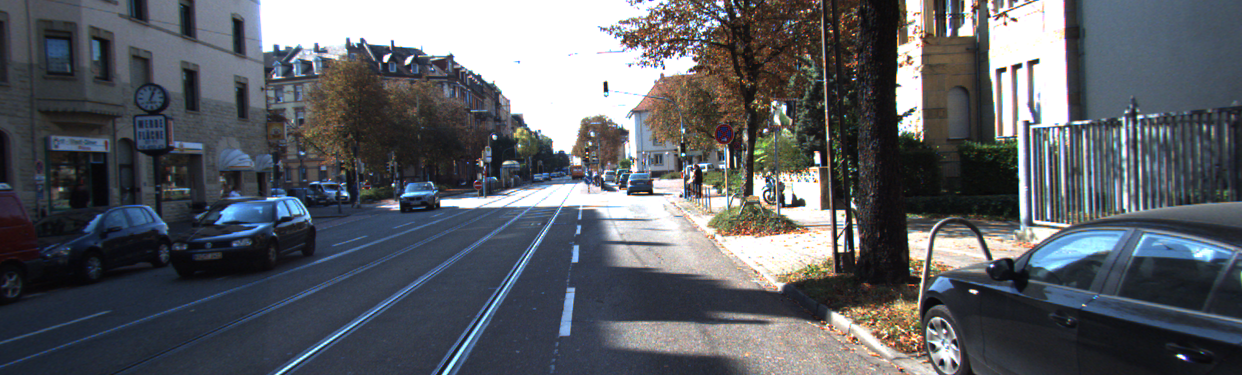}} &
\makecell*[c]{\includegraphics[width=\picwidth]{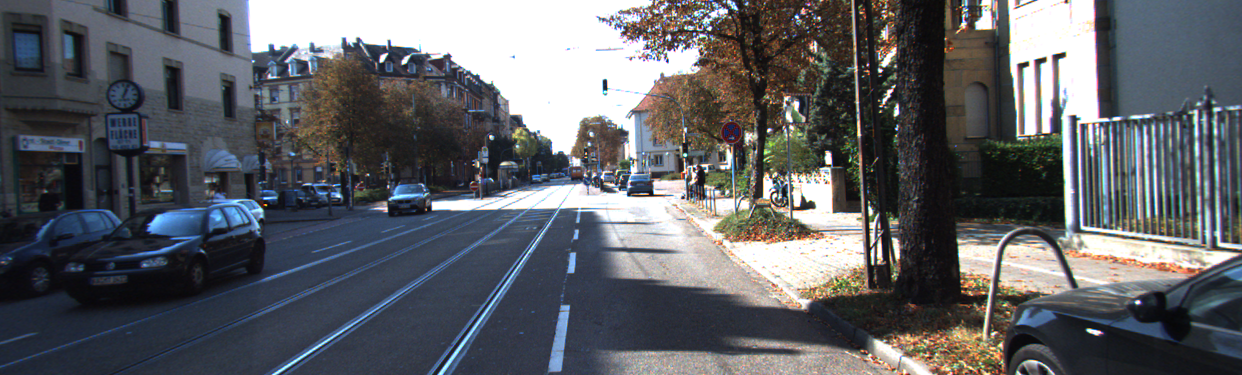}} \\

\makecell*[c]{PWC-Net\\\hline 13.87} & 
\makecell*[c]{\includegraphics[width=\picwidth]{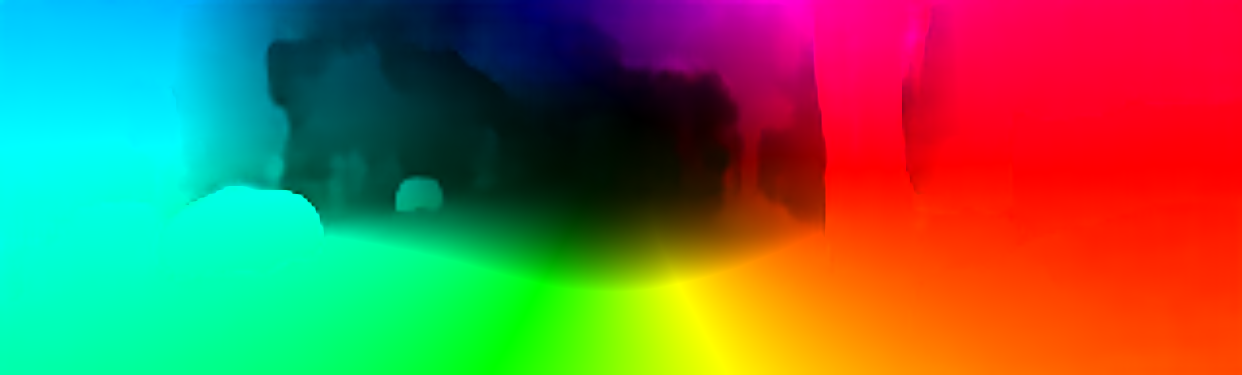}} & 
\makecell*[c]{\includegraphics[width=\picwidth]{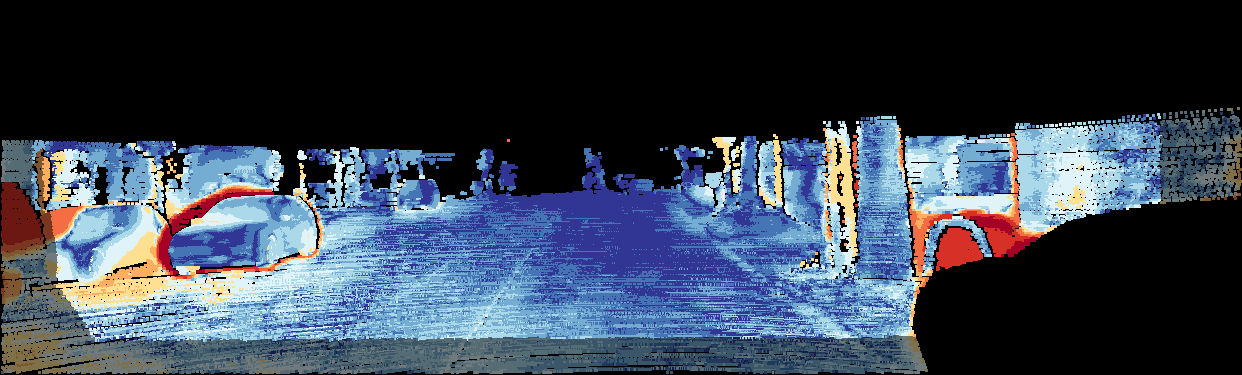}} \\

\makecell*[c]{HD$^3$\\\hline 6.79} & 
\makecell*[c]{\includegraphics[width=\picwidth]{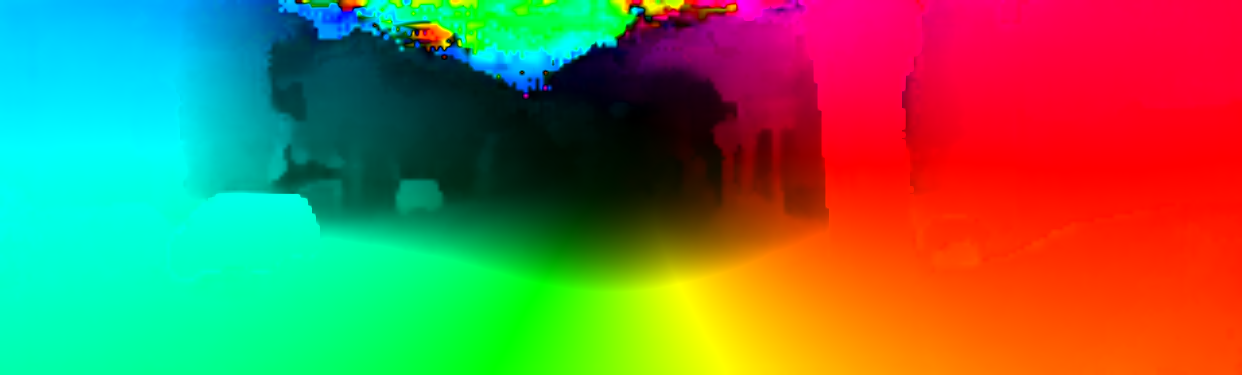}} &
\makecell*[c]{\includegraphics[width=\picwidth]{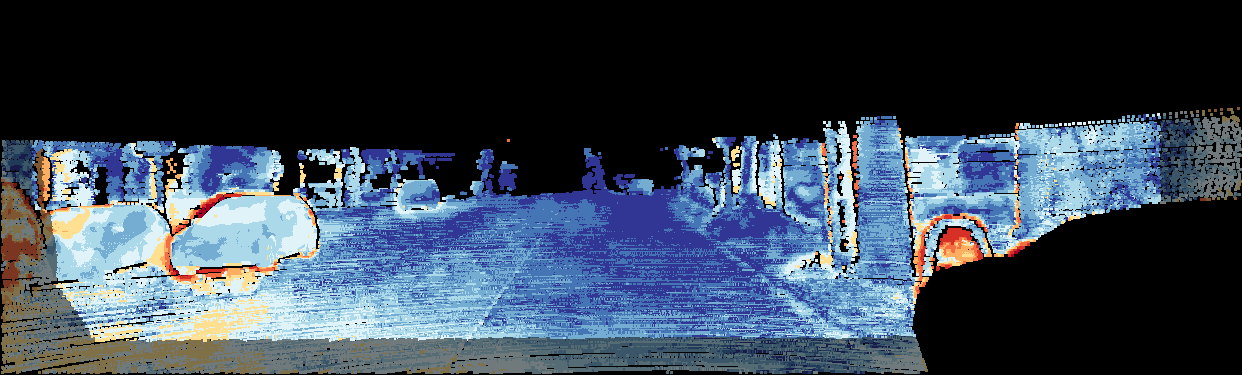}} \\

\makecell*[c]{VCN\\\hline 6.09} & 
\makecell*[c]{\includegraphics[width=\picwidth]{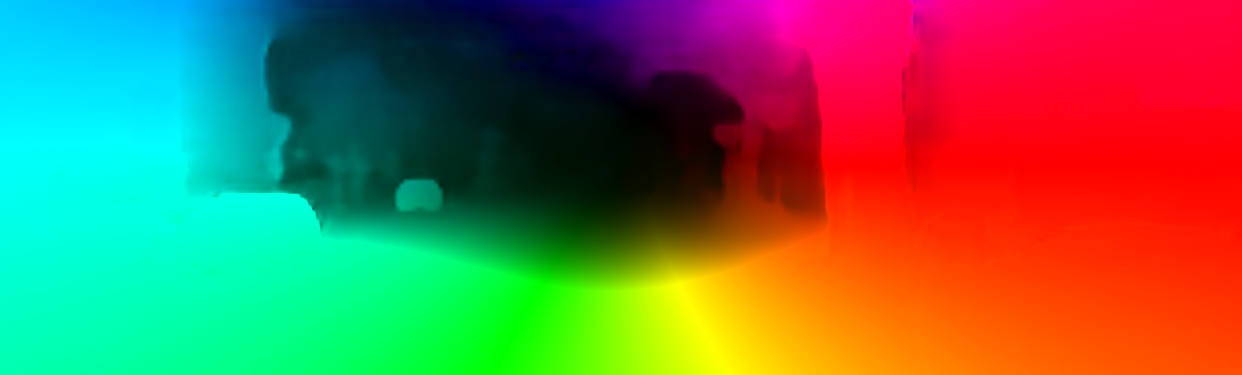}} &
\makecell*[c]{\includegraphics[width=\picwidth]{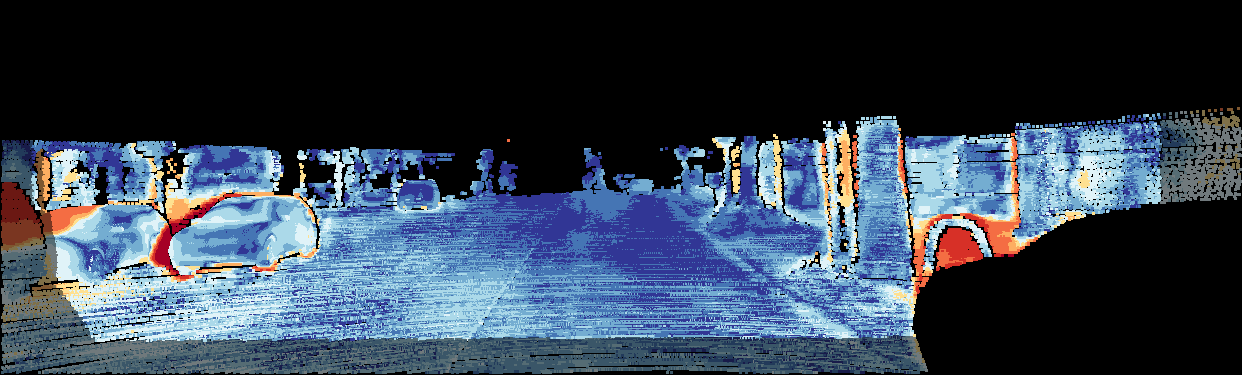}} \\

\makecell*[c]{VCN+LCV\\\hline 5.70} &
\makecell*[c]{\includegraphics[width=\picwidth]{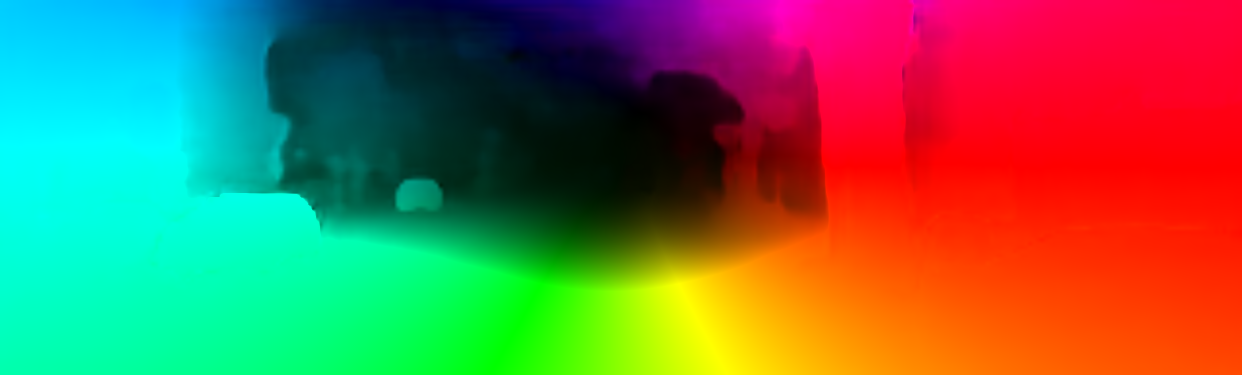}} &
\makecell*[c]{\includegraphics[width=\picwidth]{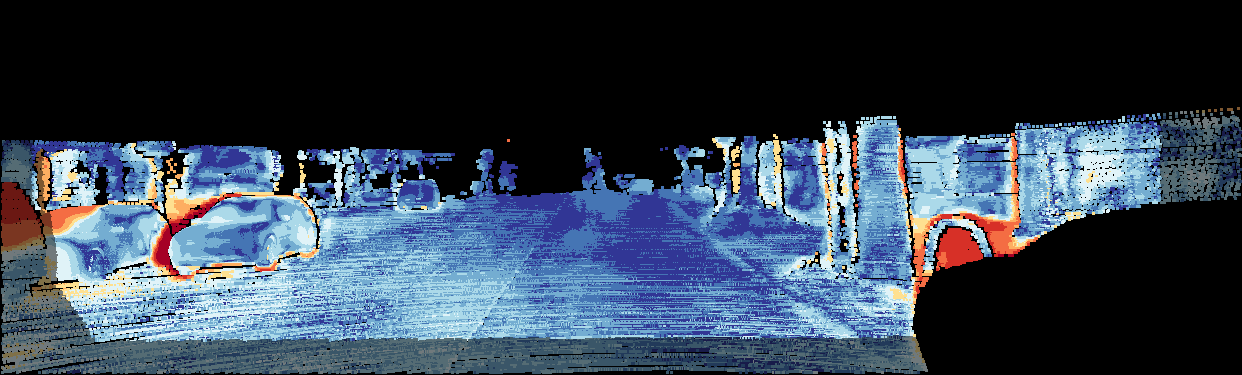}}
\end{tabular}
\caption{
    More visualization results on the KITTI 2015 test set.
    The number under each method name denotes the Fl-all score on the given frames.
    The estimated flow and error maps are presented on the left and right sides, respectively.
    From blue to red, the error of the estimated flow increases in the error map.
}
\label{fig:more-visualization-kitti15}

\end{figure}

As shown in Fig.~\ref{fig:more-visualization-sintel} and
\ref{fig:more-visualization-kitti15}, we compare our method with
other methods under the supervised settings.
We can observe that the flow boundary of the dragon in 
Fig.~\ref{fig:more-visualization-sintel}, which is predicted by VCN+LCV,
is better than the other methods and the flow prediction
near the tree (in front of the car) and fence by our method in
Fig.~\ref{fig:more-visualization-kitti15} is more accurate
compared with those of the others.
The flow prediction for these pixels are are challenging due to the occlusion. LCV explores more information among channel dimensions, which could help alleviate the problem of occlusion to some extent.

\begin{comment}

\end{comment}

\subsection{More Visualization Result on Challenging Cases}

We provide three videos showing the effectiveness of our method in three types of challenging cases: 1) illumination change; 2) noise; and 3) adversarial patches. The test videos are from the training set in the KITTI tracking benchmark. We compare the flow results under the normal setting with those under challenging settings. We can observer that the flow results of our model in either one of three challenging cases are temporarily consistent and reasonably good.

\subsection{Visualization of the Learned Features}

\begin{figure}[!htb]
\centering
\def\picwidth{0.8\textwidth}

\subfloat[frame]{\includegraphics[width=\picwidth]{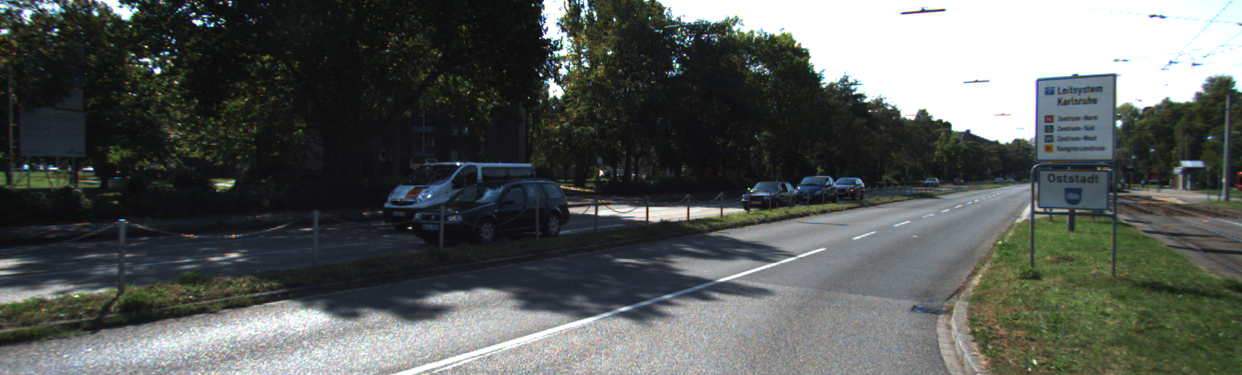}} \\
\subfloat[feature (max eigenvalue)]{\includegraphics[width=\picwidth]{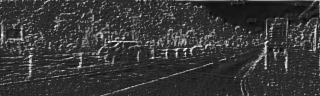}} \\
\subfloat[feature (min eigenvalue)]{\includegraphics[width=\picwidth]{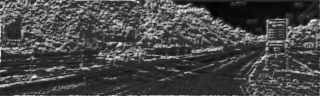}} 
\caption{The feature maps corresponding to the largest the smallest eigenvalues.}
\label{fig:visualization-feature}
%\vspace{0mm}
\end{figure}

We visualize the feature maps for different eigenvalues in Fig.~\ref{fig:visualization-feature}. We find that boundaries of (moving) objects are salient in the feature map corresponding to the max eigenvalue while the min eigenvalue mainly corresponds to background information, which results in more discriminative cost volume and more accurate flow estimation.

\end{document}